\documentclass[12pt]{article}

\usepackage[T1]{fontenc}
\usepackage[utf8]{inputenc}
\usepackage{lmodern}           

\usepackage[a4paper, margin=2.5cm]{geometry}
\usepackage{graphicx}
\usepackage{tikz}
\definecolor{DynTissue}{HTML}{EAD1CF}
\definecolor{DynOutline}{HTML}{8C5957}
\definecolor{DynGrid}{HTML}{4E7885}
\definecolor{DynThin}{HTML}{BDCDD0}
\definecolor{DynDistance}{HTML}{587EAA}
\definecolor{DynArea}{HTML}{1764A0}
\definecolor{DynTube}{HTML}{C36B28}
\usepackage{booktabs}
\usepackage{subcaption}
\usepackage[font=small]{caption}
\usepackage{placeins}

\usepackage{amsmath}
\usepackage{amssymb}

\usepackage{makecell}

\usepackage{float}

\usepackage[authoryear]{natbib}
\usepackage{hyperref}
\definecolor{ManuscriptLink}{HTML}{155CA2}
\hypersetup{colorlinks=true,linkcolor=ManuscriptLink,citecolor=ManuscriptLink,urlcolor=ManuscriptLink}

\usepackage{tipa}
\usepackage{tipx}              

\title{\textbf{Vocal-Tract Acoustics, Aerodynamics, and Overlap-and-Add Synthesis in DYNARTmo}}
\author{
  Bernd J.~Kröger$^{1,2}$\\[1ex]
  {\small $^{1}$Medical School, RWTH Aachen University, Aachen, Germany}\\
  {\small $^{2}$Kröger Lab, Belgium, \url{www.speechtrainer.eu}}\\[2ex]
}
\date{}
\begin{document}
\clubpenalty=3000
\widowpenalty=3000
\maketitle
\begingroup
\footnotesize
\noindent\copyright\ 2026 Bernd J.~Kröger. This article and its figures are licensed under
\href{https://creativecommons.org/licenses/by/4.0/}{Creative Commons Attribution 4.0 International (CC BY 4.0)}.
The accompanying Python reference package is separately licensed for noncommercial use and is not covered by the article's CC BY licence.
\par
\endgroup

\begin{abstract}
This paper presents the acoustic and aerodynamic extension of DYNARTmo, a dynamic articulatory model for visualizing and synthesizing speech production. Gestural control trajectories determine vocal-tract geometry, which is converted into an acoustic tube with nasal branching and effective lateral passages. A lumped aerodynamic network relates prescribed glottal motion and pulmonary pressure to airflow and supraglottal pressure storage. Phonatory excitation, oral friction, and glottal aspiration are synthesized as separate contributions. Frame-specific acoustic responses are combined by overlap-and-add, preserving response tails beyond individual control frames. Three German words, ``Tasse'', ``Muse'', and ``Lade'', illustrate the coordination of articulation, pressure, flow, and acoustic sources across vowels, nasals, laterals, fricatives, and plosives. The implementation has been optimized by the author primarily for sound recognizability; independent perceptual evaluation remains outstanding, and fricative-noise naturalness requires further refinement. The model provides an inspectable, articulatorily controlled production plant for future neural simulations of babbling, imitation, and early word acquisition. Further work will improve acoustic realism, reduce empirical source corrections, and extend the sensory information available to developmental learning models.
\end{abstract}

\section{Introduction}
Articulatory speech synthesis relates a specification of speech movements to the physical conditions under which sound is generated and propagated. In addition to reproducing an acoustic waveform, an articulatory model can make the relationships between vocal-tract configuration and sound accessible to inspection: how the shapes and coupling of resonant cavities determine vocalic resonances, and how constrictions influence airflow, pressure, and sound generation. This is particularly useful in an application that presents synchronized views of articulator movement and the acoustic signal. The present paper describes the acoustic and aerodynamic extension of DYNARTmo, a model originally developed for the visualization of speech articulation.

Four preceding publications describe the articulatory basis of this extension. The initial model specifies articulator configurations and their contextual combination \citep{kroeger2025geometry}; its 2D+ extension estimates tongue--palate contact from midsagittal geometry and a transverse palatal representation \citep{kroeger2025palate}. Subsequent work describes gestural movement generation \citep{kroeger2025gestures} and the coordination of primary and supporting articulators \citep{kroeger2025tradeoff}. These components provide the evolving geometry used here. The present contribution concerns the transformation of that geometry into an acoustic tract, the calculation of pressure and airflow, the construction of excitation sources, and the assembly of a synthetic signal. Articulation is represented by overlapping gestures rather than by a succession of synchronous segment-sized configurations. Acoustic segment boundaries serve as reference landmarks and must be distinguished from gesture-specific activation intervals; this distinction is developed for the examples in Section~\ref{sec:examples}.

The acoustic framework follows the source--filter perspective of \citet{fant1960} and the digital acoustic-tube tradition associated with \citet{kelly1962}. A tract is represented by connected sections whose areas determine characteristic impedances and wave scattering. Such representations support both impulse-response calculations and simulations with a time-dependent glottal boundary. The distinction between a succession of static tract configurations and a continuously changing acoustic system is important in interpreting these approaches \citep{liljencrants1985}. DYNARTmo uses responses computed for successive tract configurations and combines them by superposition; it does not propagate one uninterrupted acoustic wave state through all articulatory changes.

Aerodynamic modeling adds a different set of requirements. Glottal inflow, supraglottal pressure storage, and oral and nasal outflow must remain distinguishable, especially during closure and release. The analysis of released plosives by \citet{rothenberg1968} provides a historical basis for considering these quantities together with glottal--oral coordination. Source--tract interaction is also central to self-oscillating models such as that of \citet{ishizaka1972}. The present system adopts a more restricted approach: glottal motion and its timing are prescribed, whereas pressure and flow respond to that motion and to the tract configuration. It therefore does not derive vocal-fold oscillation or fundamental frequency from a mechanical stability calculation.

The synthesis strategy uses overlap-and-add in the general sense of adding the responses to successive excitation contributions. Fast convolution \citep{stockham1966} and pitch-synchronous waveform processing \citep{moulines1990} provide related historical contexts, but neither is identical to the procedure described here. Phonatory synthesis adds computed pulse responses at glottal pulse times; noise synthesis adds locally filtered contributions from continuous random signals. The system does not obtain its synthesis units by windowing and rearranging recorded speech. Three temporal concepts must be distinguished: \emph{control frames}, \emph{glottal pulse times}, and \emph{acoustic response duration}. Control frames are the fixed $5$-ms intervals at which articulatory and physiological control values are updated. Glottal pulse times belong to a separate timing grid derived from the frequency specified by the vocal cord tension (CT) control; in voiced segments, they determine when phonatory responses are inserted. Acoustic response duration describes how long the response to an individual excitation contributes to the output, including the subsequent ringing of the tract. This duration can exceed both a control frame and the interval between successive glottal pulses, allowing later responses to overlap with earlier ones.

A basic organizing assumption is that CT provides a continuous underlying timing control across voiced and voiceless segments. In the implementation, the CT-derived phase and its nominal glottal pulse times continue through voiceless intervals, while phonatory excitation is suppressed when the glottal configuration provides no oscillatory component. The model thus distinguishes an underlying timing process from the presence of an acoustic fundamental frequency: nominal pulse times in a voiceless interval are computational timing events, not actual vocal-fold pulses. This working hypothesis and its implementation are specified in Section~\ref{sec:ola}.

This paper documents the current implementation and its simplifying assumptions. The description distinguishes the calculation of sound propagation, the calculation of pressure and airflow, and the empirical adjustment of excitation-source spectra and amplitudes. Some source parameters are tuned by listening to the synthetic output rather than derived directly from physical relations. These adjustments can improve the resulting sound, but a plausible sound alone does not establish that the underlying pressure, airflow, and source-amplitude estimates are physiologically accurate. The model calculations and the empirically tuned parameters are therefore identified explicitly. The scope includes vowels, nasals, laterals, voiceless and voiced fricatives, and voiceless and voiced plosive transitions. Quantitative evaluation and the selection of illustrative natural--synthetic comparisons are treated separately from the model specification.

\section{From Articulatory Gestures to the Acoustic Vocal Tract}
\subsection{Control hierarchy and temporal representation}
The input to the acoustic extension is a set of articulatory trajectories generated from the gestural organization described in the preceding papers. Vocalic shape formation, consonantal constriction, velopharyngeal control, and glottal control jointly determine the articulatory configuration. Additional trajectories specify pulmonary pressure and cord tension (CT). These commands are updated at the boundaries of successive control frames, with the fixed control-frame duration
\begin{equation}
 t_k=k\Delta_c, \qquad \Delta_c=5\,\mathrm{ms}.
 \label{eq:controlgrid}
\end{equation}
Each \emph{control frame} spans one interval $\Delta_c$ and has an associated articulatory and physiological configuration. It is neither an analysis window applied to the final waveform nor a glottal period.

For each control frame, the articulator configuration is transformed into an area function, acoustic response kernels, and aerodynamic quantities. Their acoustic response duration can extend across several control frames. During utterance preparation, configurations are processed chronologically so that the slowly varying pressure state can be transferred to the next frame. Glottal pulse times are derived separately from CT and the selected segmental frequency adjustment (Section~\ref{sec:ola}). The interval between consecutive glottal pulse times can therefore span several control frames; the two timing grids are not required to coincide.

The implementation retains the grouping of gestures belonging to the same segment. This information supplies a context for selected empirical controls, including the treatment of voiced fricatives and the additional calibration of voiceless-plosive friction. For example, a vowel's anticipatory phonation gesture must not cause a preceding voiceless oral constriction to receive the voiced-fricative modulation. This use of gesture grouping is distinguished from the generation of aerodynamic drive: the current continuous oral-noise calculation does not require a phoneme label or a plosive-release event to enable a source.

\subsection{Area functions and tract topology}
\begin{figure}[p]
\centering
\begin{subfigure}[t]{0.48\textwidth}\centering
\resizebox{\linewidth}{!}{\begin{tikzpicture}[x=0.018cm,y=-0.018cm,line cap=round,line join=round]
\path[use as bounding box] (202,77) rectangle (481,382);
\clip (202,77) rectangle (481,382);
\fill[fill=DynTissue] (380.5500,297.4500) -- (381.7000,286.6500) -- (380.8000,274.8500) -- (380.0500,266.0500) -- (380.3500,255.2500) -- (383.5000,245.6000) -- (383.7500,237.7000) -- (380.0000,231.8000) -- (375.9500,238.4500) -- (373.6000,245.3500) -- (372.3000,254.1000) -- (369.0000,262.8500) -- (365.5000,272.5000) -- (355.9000,252.4500) -- (363.1000,247.6000) -- (368.4500,242.6500) -- (370.8500,232.8500) -- (366.7000,222.0500) -- (369.6500,213.0500) -- (373.3500,203.9500) -- (377.0000,192.1000) -- (376.9000,182.1000) -- (374.8000,169.2000) -- (371.7500,159.2000) -- (366.7500,147.3500) -- (359.7000,135.6000) -- (352.4000,128.9000) -- (342.2000,123.3000) -- (332.1000,120.6000) -- (321.0000,121.7000) -- (310.9000,127.5000) -- (304.6500,133.3500) -- (297.4500,138.2000) -- (291.3000,145.0000) -- (283.3000,150.9000) -- (277.4500,155.9000) -- (280.3500,162.8000) -- (284.2500,168.7500) -- (287.3000,173.8000) -- (287.8250,191.2500) -- (250.3500,197.2000) -- (244.3500,200.2000) -- (236.3500,205.2000) -- (229.8750,212.2000) -- (226.8250,221.2000) -- (225.4000,229.2500) -- (226.8250,238.2000) -- (230.8750,243.3000) -- (236.8250,248.2500) -- (243.8750,252.2000) -- (252.3500,255.1500) -- (260.3000,258.1000) -- (267.9250,262.0000) -- (276.0750,266.8000) -- (286.0250,272.6000) -- (295.1250,280.2500) -- (303.3250,286.0500) -- (311.6500,294.8000) -- (318.4000,301.0500) -- (322.7250,312.1500) -- (324.7750,324.8000) -- (327.1250,337.8000) -- (332.4500,346.8500) -- (336.8250,354.1500) -- (340.6750,369.0000) -- (342.1000,381.0000) -- (369.1500,379.0500) -- (362.1000,343.5500) -- (407.8500,332.7500) -- (390.9000,328.7000) -- (357.3000,336.5500) -- (359.3000,318.5000) -- (380.5500,297.4500) -- cycle;
\fill[fill=DynTissue] (265.0000,148.0000) -- (276.0000,137.0000) -- (280.0000,132.0000) -- (286.0000,127.0000) -- (293.0000,124.0000) -- (300.0000,120.0000) -- (306.0000,115.0000) -- (312.0000,110.0000) -- (325.0000,113.0000) -- (317.0000,115.0000) -- (311.0000,120.0000) -- (305.0000,124.0000) -- (300.0000,129.0000) -- (294.0000,133.0000) -- (289.0000,136.0000) -- (277.0000,136.0000) -- cycle;
\fill[fill=DynTissue] (250.0000,94.0000) -- (249.0000,104.0000) -- (246.0000,112.0000) -- (243.0000,121.0000) -- (243.0000,128.0000) -- (248.0000,134.0500) -- (253.0000,136.1000) -- (260.0000,136.1000) -- (265.0000,134.1000) -- (269.0000,130.0000) -- (273.0000,125.0000) -- (277.0000,120.0000) -- (282.0000,115.0000) -- (286.0000,108.0000) -- (289.0000,103.0000) -- (292.0000,97.0000) -- cycle;
\fill[fill=DynTissue] (250.3500,197.2000) -- (254.3000,193.2500) -- (251.2500,189.2500) -- (247.2500,184.2500) -- (245.2000,177.2500) -- (243.2000,171.2500) -- (245.1500,165.2500) -- (248.2000,160.2500) -- (253.2500,157.1500) -- (259.2500,156.1000) -- (266.2500,160.0500) -- (270.2500,165.2000) -- (272.2500,173.2500) -- (273.2500,184.2500) -- (276.2500,194.2500) -- (272.2500,204.2500) -- cycle;
\fill[fill=DynTissue] (325.0000,113.0000) -- (333.0000,113.0000) -- (344.0000,116.0000) -- (353.9450,119.9000) -- (363.0450,123.8550) -- (373.2450,128.9100) -- (382.3450,133.7100) -- (392.6350,136.5750) -- (402.9700,139.2300) -- (410.0700,141.6850) -- (415.2800,145.3950) -- (417.1250,153.4500) -- (417.0250,161.3500) -- (416.8250,169.2500) -- (418.8900,178.3300) -- (425.0900,181.1650) -- (429.9350,179.0100) -- (433.9800,174.1650) -- (434.9350,167.1200) -- (436.0350,158.0400) -- (437.1350,151.0400) -- (437.2450,144.9850) -- (438.3900,135.9950) -- (438.3800,128.8500) -- (434.2250,123.0050) -- (427.0800,120.1950) -- (416.5900,120.6850) -- (406.1000,121.0750) -- (398.0550,120.3650) -- (388.9450,118.4100) -- (379.9450,118.5100) -- (372.2000,119.7550) -- (364.0000,117.9000) -- (357.1000,115.0000) -- cycle;
\draw[DynOutline,line width=0.42pt] (265.0000,148.0000) -- (267.0000,136.0000) -- (270.0000,130.0000) -- (275.0000,122.0000) -- (280.0000,117.0000) -- (285.0000,110.0000) -- (289.0000,103.0000) -- (293.0000,96.0000);
\draw[DynOutline,line width=0.42pt] (265.0000,148.0000) -- (276.0000,137.0000) -- (280.0000,132.0000) -- (286.0000,127.0000) -- (293.0000,124.0000) -- (300.0000,120.0000) -- (306.0000,115.0000) -- (312.0000,110.0000);
\draw[DynOutline,line width=0.42pt] (277.0000,136.0000) -- (289.0000,136.0000) -- (294.0000,133.0000) -- (300.0000,129.0000) -- (305.0000,124.0000) -- (311.0000,120.0000) -- (317.0000,115.0000) -- (325.0000,113.0000);
\draw[DynOutline,line width=0.42pt] (250.0000,93.0000) -- (250.0000,86.0000) -- (248.0000,79.0000) -- (242.0000,74.0000) -- (235.0000,70.0000) -- (230.0000,66.0000) -- (227.0000,60.0000) -- (228.0000,53.0000);
\draw[DynOutline,line width=0.42pt] (465.0000,378.0000) -- (463.0000,370.0000) -- (463.0000,364.0000) -- (462.0000,357.0000) -- (462.0000,349.0000) -- (461.0000,339.0000) -- (461.0000,329.0000) -- (462.0000,322.0000) -- (463.0000,312.0000) -- (463.0000,304.0000) -- (464.0000,294.0000) -- (465.0000,285.0000) -- (466.0000,277.0000) -- (467.0000,268.0000) -- (469.0000,257.0000) -- (470.0000,247.0000) -- (471.0000,235.0000) -- (473.0000,226.0000) -- (474.0000,218.0000) -- (476.0000,209.0000) -- (477.0000,203.0000) -- (478.0000,194.0000) -- (479.0000,187.0000) -- (480.0000,180.0000);
\draw[DynOutline,line width=0.42pt] (432.0000,379.0000) -- (432.0000,371.0000) -- (430.0000,361.0000) -- (431.0000,354.0000) -- (431.0000,347.0000) -- (431.0000,339.0000) -- (431.0000,332.0000) -- (430.0000,322.0000) -- (431.0000,314.0000) -- (431.0000,302.0000) -- (432.0000,293.0000) -- (434.0000,282.0000) -- (434.0000,272.0000) -- (435.0000,264.0000) -- (436.0000,255.0000) -- (438.0000,245.0000) -- (439.0000,235.0000) -- (441.0000,226.0000) -- (443.0000,219.0000) -- (446.0000,212.0000) -- (449.0000,204.0000) -- (453.0000,196.0000) -- (456.0000,190.0000) -- (458.0000,182.0000);
\draw[DynOutline,line width=0.42pt] (458.0000,182.0000) -- (460.0000,177.0000) -- (463.0000,173.0000) -- (468.0000,171.0000) -- (473.0000,171.0000) -- (478.0000,175.0000) -- (480.0000,180.0000);
\draw[DynOutline,line width=0.42pt] (438.0000,245.0000) -- (443.0000,242.0000) -- (451.0000,239.0000) -- (457.0000,240.0000) -- (464.0000,241.0000) -- (469.0000,243.0000);
\draw[DynOutline,line width=0.42pt] (437.0000,251.0000) -- (442.0000,250.0000) -- (448.0000,250.0000) -- (456.0000,250.0000) -- (463.0000,251.0000) -- (469.0000,253.0000);
\draw[DynOutline,line width=0.42pt] (433.0000,280.0000) -- (441.0000,279.0000) -- (446.0000,278.0000) -- (453.0000,279.0000) -- (459.0000,280.0000) -- (465.0000,283.0000);
\draw[DynOutline,line width=0.42pt] (433.0000,290.0000) -- (438.0000,290.0000) -- (445.0000,290.0000) -- (451.0000,290.0000) -- (457.0000,290.0000) -- (463.0000,292.0000);
\draw[DynOutline,line width=0.42pt] (431.0000,320.0000) -- (436.0000,317.0000) -- (443.0000,316.0000) -- (449.0000,316.0000) -- (456.0000,317.0000) -- (461.0000,319.0000);
\draw[DynOutline,line width=0.42pt] (431.0000,331.0000) -- (437.0000,329.0000) -- (443.0000,327.0000) -- (450.0000,327.0000) -- (456.0000,327.0000) -- (461.0000,328.0000);
\draw[DynOutline,line width=0.42pt] (431.0000,355.0000) -- (436.0000,353.0000) -- (443.0000,352.0000) -- (450.0000,352.0000) -- (456.0000,352.0000) -- (461.0000,354.0000);
\draw[DynOutline,line width=0.42pt] (432.0000,365.0000) -- (438.0000,363.0000) -- (445.0000,362.0000) -- (453.0000,362.0000) -- (458.0000,362.0000) -- (463.0000,364.0000);
\draw[DynOutline,line width=0.42pt] (272.3000,164.2000) -- (269.8250,173.2000) -- (272.3000,183.2000) -- (274.8250,191.2500) -- (273.3500,199.2500) -- (270.3500,206.2000) -- (264.8250,214.2500) -- (261.3000,223.2000) -- (260.3500,231.2500) -- (265.8250,235.2000);
\draw[DynOutline,line width=0.42pt] (272.3000,164.2000) -- (278.3500,173.2000) -- (284.3000,181.2500) -- (287.8250,191.2500) -- (287.8250,200.2000) -- (288.3500,210.2000) -- (286.3500,219.2500) -- (281.3500,227.2000) -- (273.8250,233.2000) -- (265.3500,235.2000);
\draw[DynOutline,line width=0.42pt] (250.3500,197.2000) -- (244.3500,200.2000) -- (236.3500,205.2000) -- (229.8750,212.2000) -- (226.8250,221.2000) -- (225.4000,229.2500) -- (226.8250,238.2000) -- (230.8750,243.3000) -- (236.8250,248.2500) -- (243.8750,252.2000) -- (252.3500,255.1500) -- (260.3000,258.1000) -- (267.9250,262.0000) -- (276.0750,266.8000) -- (286.0250,272.6000) -- (295.1250,280.2500) -- (303.3250,286.0500) -- (311.6500,294.8000) -- (318.4000,301.0500) -- (322.7250,312.1500) -- (324.7750,324.8000) -- (327.1250,337.8000) -- (332.4500,346.8500) -- (336.8250,354.1500) -- (340.6750,369.0000) -- (342.1000,381.0000);
\draw[DynOutline,line width=0.42pt] (407.8500,379.0500) -- (407.8500,332.7500) -- (362.1000,343.5500) -- (369.1500,379.0500);
\draw[DynOutline,line width=0.42pt] (438.0500,122.9500) -- (423.8500,301.3500) -- (393.2500,298.4000) -- (388.1000,317.1000) -- (390.9000,328.7000) -- (357.3000,336.5500) -- (359.3000,318.5000) -- (380.5500,297.4500);
\draw[DynOutline,line width=0.42pt] (250.0000,94.0000) -- (249.0000,104.0000) -- (246.0000,112.0000) -- (243.0000,121.0000) -- (243.0000,128.0000) -- (248.0000,134.0500) -- (253.0000,136.1000) -- (260.0000,136.1000) -- (265.0000,134.1000) -- (269.0000,130.0000) -- (273.0000,125.0000) -- (277.0000,120.0000) -- (282.0000,115.0000) -- (286.0000,108.0000) -- (289.0000,103.0000) -- (292.0000,97.0000);
\draw[DynOutline,line width=0.42pt] (250.3500,197.2000) -- (254.3000,193.2500) -- (251.2500,189.2500) -- (247.2500,184.2500) -- (245.2000,177.2500) -- (243.2000,171.2500) -- (245.1500,165.2500) -- (248.2000,160.2500) -- (253.2500,157.1500) -- (259.2500,156.1000) -- (266.2500,160.0500) -- (270.2500,165.2000) -- (272.2500,173.2500) -- (273.2500,184.2500) -- (276.2500,194.2500) -- (272.2500,204.2500);
\draw[DynOutline,line width=0.42pt] (325.0000,113.0000) -- (333.0000,113.0000) -- (344.0000,116.0000) -- (353.9450,119.9000) -- (363.0450,123.8550) -- (373.2450,128.9100) -- (382.3450,133.7100) -- (392.6350,136.5750) -- (402.9700,139.2300) -- (410.0700,141.6850) -- (415.2800,145.3950) -- (417.1250,153.4500) -- (417.0250,161.3500) -- (416.8250,169.2500) -- (418.8900,178.3300) -- (425.0900,181.1650) -- (429.9350,179.0100) -- (433.9800,174.1650) -- (434.9350,167.1200) -- (436.0350,158.0400) -- (437.1350,151.0400) -- (437.2450,144.9850) -- (438.3900,135.9950) -- (438.3800,128.8500) -- (434.2250,123.0050) -- (427.0800,120.1950) -- (416.5900,120.6850) -- (406.1000,121.0750) -- (398.0550,120.3650) -- (388.9450,118.4100) -- (379.9450,118.5100) -- (372.2000,119.7550) -- (364.0000,117.9000) -- (357.1000,115.0000);
\draw[DynOutline,line width=0.42pt] (380.5500,297.4500) -- (381.7000,286.6500) -- (380.8000,274.8500) -- (380.0500,266.0500) -- (380.3500,255.2500) -- (383.5000,245.6000) -- (383.7500,237.7000) -- (380.0000,231.8000) -- (375.9500,238.4500) -- (373.6000,245.3500) -- (372.3000,254.1000) -- (369.0000,262.8500) -- (365.5000,272.5000) -- (355.9000,252.4500) -- (363.1000,247.6000) -- (368.4500,242.6500) -- (370.8500,232.8500) -- (366.7000,222.0500) -- (369.6500,213.0500) -- (373.3500,203.9500) -- (377.0000,192.1000) -- (376.9000,182.1000) -- (374.8000,169.2000) -- (371.7500,159.2000) -- (366.7500,147.3500) -- (359.7000,135.6000) -- (352.4000,128.9000) -- (342.2000,123.3000) -- (332.1000,120.6000) -- (321.0000,121.7000) -- (310.9000,127.5000) -- (304.6500,133.3500) -- (297.4500,138.2000) -- (291.3000,145.0000) -- (283.3000,150.9000) -- (277.4500,155.9000) -- (280.3500,162.8000) -- (284.2500,168.7500) -- (287.3000,173.8000) -- (287.8250,191.2500);
\draw[DynThin,line width=0.40pt] (350.0000,200.0000) -- (431.9171,200.0000);
\draw[DynThin,line width=0.40pt] (350.0000,200.0000) -- (425.9643,184.8898);
\draw[DynThin,line width=0.40pt] (350.0000,200.0000) -- (420.4250,170.8290);
\draw[DynThin,line width=0.40pt] (350.0000,200.0000) -- (414.8472,156.6705);
\draw[DynThin,line width=0.40pt] (350.0000,200.0000) -- (408.7659,141.2341);
\draw[DynThin,line width=0.40pt] (350.0000,200.0000) -- (392.4193,136.5150);
\draw[DynThin,line width=0.40pt] (350.0000,200.0000) -- (378.3345,131.5946);
\draw[DynThin,line width=0.40pt] (350.0000,200.0000) -- (364.9576,124.8029);
\draw[DynThin,line width=0.40pt] (350.0000,200.0000) -- (350.0000,118.3529);
\draw[DynThin,line width=0.40pt] (350.0000,200.0000) -- (336.3659,113.9180);
\draw[DynThin,line width=0.40pt] (350.0000,200.0000) -- (322.1887,114.4056);
\draw[DynThin,line width=0.40pt] (350.0000,200.0000) -- (309.6847,120.8768);
\draw[DynThin,line width=0.40pt] (350.0000,200.0000) -- (298.9325,129.7117);
\draw[DynThin,line width=0.40pt] (350.0000,200.0000) -- (291.4383,134.5370);
\draw[DynThin,line width=0.40pt] (275.0000,125.0000) -- (290.1481,145.8495);
\draw[DynThin,line width=0.40pt] (275.0000,125.0000) -- (286.8592,148.2751);
\draw[DynThin,line width=0.40pt] (275.0000,125.0000) -- (283.3931,150.8313);
\draw[DynThin,line width=0.40pt] (275.0000,125.0000) -- (279.6027,154.0601);
\draw[DynThin,line width=0.40pt] (275.0000,125.0000) -- (275.0000,168.2165);
\draw[DynGrid,line width=0.65pt] (371.3131,306.6000) -- (390.9917,306.6000);
\draw[DynGrid,line width=0.65pt] (362.5556,315.2750) -- (388.6026,315.2750);
\draw[DynGrid,line width=0.65pt] (358.6961,323.9500) -- (389.7534,323.9500);
\draw[DynGrid,line width=0.65pt] (357.3000,336.5500) -- (390.9000,328.7000);
\draw[DynGrid,line width=0.65pt] (370.5321,210.8806) -- (431.0510,210.8806);
\draw[DynGrid,line width=0.65pt] (374.4585,221.7611) -- (430.1850,221.7611);
\draw[DynGrid,line width=0.65pt] (380.0261,232.6417) -- (429.3189,232.6417);
\draw[DynGrid,line width=0.65pt] (380.3633,243.5222) -- (428.4529,243.5222);
\draw[DynGrid,line width=0.65pt] (380.7005,254.4028) -- (427.5868,254.4028);
\draw[DynGrid,line width=0.65pt] (381.0378,265.2833) -- (426.7208,265.2833);
\draw[DynGrid,line width=0.65pt] (381.3750,276.1639) -- (425.8547,276.1639);
\draw[DynGrid,line width=0.65pt] (381.6580,287.0444) -- (424.9887,287.0444);
\draw[DynGrid,line width=0.65pt] (380.5500,297.4500) -- (393.2500,298.4000);
\draw[DynGrid,line width=0.65pt] (268.6000,163.0756) -- (268.6000,130.4100);
\draw[DynGrid,line width=0.65pt] (262.2000,157.7646) -- (262.2000,135.2200);
\draw[DynGrid,line width=0.65pt] (255.8000,156.7038) -- (255.8000,136.1000);
\draw[DynGrid,line width=0.65pt] (249.4000,159.5134) -- (249.4000,134.6240);
\draw[DynGrid,line width=0.65pt] (245.1500,165.2500) -- (243.0000,128.0000);
\draw[DynGrid,line width=0.65pt] (374.5667,200.0000) -- (431.9171,200.0000);
\draw[DynGrid,line width=0.65pt] (376.1701,194.7945) -- (425.9643,184.8898);
\draw[DynGrid,line width=0.65pt] (376.9673,188.8298) -- (420.4250,170.8290);
\draw[DynGrid,line width=0.65pt] (376.8891,182.0333) -- (414.8472,156.6705);
\draw[DynGrid,line width=0.65pt] (375.6400,174.3600) -- (408.7659,141.2341);
\draw[DynGrid,line width=0.65pt] (373.4774,164.8636) -- (392.4193,136.5150);
\draw[DynGrid,line width=0.65pt] (369.3025,153.3995) -- (378.3345,131.5946);
\draw[DynGrid,line width=0.65pt] (362.0356,139.4927) -- (364.9576,124.8029);
\draw[DynGrid,line width=0.65pt] (350.0000,127.5824) -- (350.0000,118.3529);
\draw[DynGrid,line width=0.65pt] (337.6597,122.0862) -- (336.3659,113.9180);
\draw[DynGrid,line width=0.65pt] (324.4478,121.3583) -- (322.1887,114.4056);
\draw[DynGrid,line width=0.65pt] (312.5706,126.5407) -- (309.6847,120.8768);
\draw[DynGrid,line width=0.65pt] (302.5861,134.7403) -- (298.9325,129.7117);
\draw[DynGrid,line width=0.65pt] (295.5921,140.2542) -- (291.4383,134.5370);
\draw[DynGrid,line width=0.65pt] (290.1481,145.8495) -- (282.9920,136.0000);
\draw[DynGrid,line width=0.65pt] (286.8592,148.2751) -- (280.6048,136.0000);
\draw[DynGrid,line width=0.65pt] (283.3931,150.8313) -- (278.5741,136.0000);
\draw[DynGrid,line width=0.65pt] (279.6027,154.0601) -- (276.7775,136.2225);
\draw[DynGrid,line width=0.65pt] (275.0000,168.2165) -- (275.0000,138.0000);
\fill[DynOutline] (350.0000,200.0000) circle[radius=1.65];
\node[font=\scriptsize,anchor=west] at (358.0000,207.0000) {P1};
\fill[DynOutline] (275.0000,125.0000) circle[radius=1.65];
\node[font=\scriptsize,anchor=west] at (283.0000,132.0000) {P2};
\fill[DynOutline] (374.1000,332.6250) circle[radius=1.65];
\node[font=\scriptsize,anchor=west] at (382.1000,339.6250) {G};
\fill[DynOutline] (390.9000,328.7000) circle[radius=1.65];
\fill[DynOutline] (357.3000,336.5500) circle[radius=1.65];
\fill[DynOutline] (243.0000,128.0000) circle[radius=1.65];
\fill[DynOutline] (245.1500,165.2500) circle[radius=1.65];
\end{tikzpicture}}
\caption{/i/}
\end{subfigure}
\hfill
\begin{subfigure}[t]{0.48\textwidth}\centering
\resizebox{\linewidth}{!}{\begin{tikzpicture}[x=0.018cm,y=-0.018cm,line cap=round,line join=round]
\path[use as bounding box] (202,77) rectangle (481,382);
\clip (202,77) rectangle (481,382);
\fill[fill=DynTissue] (379.6000,305.0500) -- (379.8000,295.2000) -- (380.8000,286.2500) -- (381.9500,275.5500) -- (386.0500,266.6500) -- (391.1000,257.9500) -- (391.3500,248.1500) -- (387.6000,245.1000) -- (383.5500,247.9500) -- (381.2000,255.8000) -- (377.0500,264.5500) -- (373.7500,273.3000) -- (366.4500,280.1000) -- (361.6000,257.2000) -- (368.8000,255.2000) -- (375.1000,250.2500) -- (379.4000,244.2500) -- (380.0000,235.3500) -- (384.8500,227.3000) -- (392.3500,219.1500) -- (399.8000,212.0500) -- (404.4500,203.0000) -- (408.0500,192.9500) -- (409.7500,182.0000) -- (407.6000,169.2000) -- (403.4000,158.4000) -- (395.1500,149.8000) -- (384.9500,145.1500) -- (375.8000,140.5500) -- (362.8000,138.8000) -- (348.9000,138.9000) -- (336.9500,142.8500) -- (327.8500,147.7000) -- (319.8000,152.6000) -- (313.7000,159.4500) -- (312.6000,167.3000) -- (315.5000,175.1500) -- (318.4500,183.0000) -- (315.8000,188.0500) -- (287.8250,182.7000) -- (250.3500,188.6500) -- (244.3500,190.7000) -- (236.3500,195.7000) -- (229.8750,201.7500) -- (226.8250,211.7000) -- (225.4000,219.7500) -- (226.8250,228.7000) -- (230.8750,234.7500) -- (236.8250,240.6500) -- (243.8750,243.6500) -- (252.3500,246.6000) -- (260.3000,250.5000) -- (267.9250,257.2500) -- (276.0750,263.0000) -- (286.0250,270.7000) -- (295.1250,276.4500) -- (303.3250,282.2500) -- (311.6500,290.0500) -- (318.4000,300.1000) -- (322.7250,311.2000) -- (324.7750,321.9500) -- (327.1250,330.2000) -- (332.4500,340.2000) -- (336.8250,348.4500) -- (340.6750,360.4500) -- (342.1000,379.1000) -- (359.6500,380.0000) -- (354.5000,355.9000) -- (402.1500,348.9000) -- (383.3000,345.8000) -- (354.4500,348.9000) -- (357.4000,328.0000) -- (379.6000,305.0500) -- cycle;
\fill[fill=DynTissue] (265.0000,148.0000) -- (276.0000,137.0000) -- (280.0000,132.0000) -- (286.0000,127.0000) -- (293.0000,124.0000) -- (300.0000,120.0000) -- (306.0000,115.0000) -- (312.0000,110.0000) -- (325.0000,113.0000) -- (317.0000,115.0000) -- (311.0000,120.0000) -- (305.0000,124.0000) -- (300.0000,129.0000) -- (294.0000,133.0000) -- (289.0000,136.0000) -- (277.0000,136.0000) -- cycle;
\fill[fill=DynTissue] (250.0000,94.0000) -- (244.2500,101.1500) -- (236.5000,106.3000) -- (228.7500,113.4000) -- (224.0000,125.1500) -- (226.1500,134.0500) -- (232.1000,140.8500) -- (241.0000,144.6500) -- (248.8500,145.5000) -- (255.7000,143.3000) -- (260.6500,138.3000) -- (268.4500,130.4500) -- (273.4500,123.5500) -- (280.3000,115.6000) -- (287.1000,107.7500) -- (292.0000,97.0000) -- cycle;
\fill[fill=DynTissue] (250.3500,188.6500) -- (251.4500,193.2500) -- (246.5000,189.2500) -- (235.8500,184.2500) -- (228.1000,179.1500) -- (222.3000,170.3000) -- (222.3500,162.4000) -- (225.4000,153.6000) -- (232.3500,148.6000) -- (244.0500,149.4500) -- (253.9000,152.4500) -- (262.6500,158.5500) -- (266.5500,169.4500) -- (270.4000,184.2500) -- (271.5000,198.0500) -- (267.5000,208.0500) -- cycle;
\fill[fill=DynTissue] (325.0000,113.0000) -- (333.0000,113.0000) -- (344.0000,116.0000) -- (353.9450,120.9450) -- (363.0450,125.9450) -- (373.2450,132.0450) -- (382.3450,136.8450) -- (392.6350,139.7100) -- (402.9700,141.3200) -- (410.0700,142.7300) -- (415.2800,145.3950) -- (417.1250,153.4500) -- (417.0250,161.3500) -- (416.8250,169.2500) -- (418.8900,178.3300) -- (425.0900,181.1650) -- (429.9350,179.0100) -- (433.9800,174.1650) -- (434.9350,167.1200) -- (436.0350,158.0400) -- (437.1350,151.0400) -- (437.2450,144.9850) -- (438.3900,135.9950) -- (438.3800,128.8500) -- (434.2250,125.0950) -- (427.0800,123.3300) -- (416.5900,123.8200) -- (406.1000,124.2100) -- (398.0550,124.5450) -- (388.9450,121.5450) -- (379.9450,121.6450) -- (372.2000,121.8450) -- (364.0000,117.9000) -- (357.1000,115.0000) -- cycle;
\draw[DynOutline,line width=0.42pt] (265.0000,148.0000) -- (267.0000,136.0000) -- (270.0000,130.0000) -- (275.0000,122.0000) -- (280.0000,117.0000) -- (285.0000,110.0000) -- (289.0000,103.0000) -- (293.0000,96.0000);
\draw[DynOutline,line width=0.42pt] (265.0000,148.0000) -- (276.0000,137.0000) -- (280.0000,132.0000) -- (286.0000,127.0000) -- (293.0000,124.0000) -- (300.0000,120.0000) -- (306.0000,115.0000) -- (312.0000,110.0000);
\draw[DynOutline,line width=0.42pt] (277.0000,136.0000) -- (289.0000,136.0000) -- (294.0000,133.0000) -- (300.0000,129.0000) -- (305.0000,124.0000) -- (311.0000,120.0000) -- (317.0000,115.0000) -- (325.0000,113.0000);
\draw[DynOutline,line width=0.42pt] (250.0000,93.0000) -- (250.0000,86.0000) -- (248.0000,79.0000) -- (242.0000,74.0000) -- (235.0000,70.0000) -- (230.0000,66.0000) -- (227.0000,60.0000) -- (228.0000,53.0000);
\draw[DynOutline,line width=0.42pt] (465.0000,378.0000) -- (463.0000,370.0000) -- (463.0000,364.0000) -- (462.0000,357.0000) -- (462.0000,349.0000) -- (461.0000,339.0000) -- (461.0000,329.0000) -- (462.0000,322.0000) -- (463.0000,312.0000) -- (463.0000,304.0000) -- (464.0000,294.0000) -- (465.0000,285.0000) -- (466.0000,277.0000) -- (467.0000,268.0000) -- (469.0000,257.0000) -- (470.0000,247.0000) -- (471.0000,235.0000) -- (473.0000,226.0000) -- (474.0000,218.0000) -- (476.0000,209.0000) -- (477.0000,203.0000) -- (478.0000,194.0000) -- (479.0000,187.0000) -- (480.0000,180.0000);
\draw[DynOutline,line width=0.42pt] (432.0000,379.0000) -- (432.0000,371.0000) -- (430.0000,361.0000) -- (431.0000,354.0000) -- (431.0000,347.0000) -- (431.0000,339.0000) -- (431.0000,332.0000) -- (430.0000,322.0000) -- (431.0000,314.0000) -- (431.0000,302.0000) -- (432.0000,293.0000) -- (434.0000,282.0000) -- (434.0000,272.0000) -- (435.0000,264.0000) -- (436.0000,255.0000) -- (438.0000,245.0000) -- (439.0000,235.0000) -- (441.0000,226.0000) -- (443.0000,219.0000) -- (446.0000,212.0000) -- (449.0000,204.0000) -- (453.0000,196.0000) -- (456.0000,190.0000) -- (458.0000,182.0000);
\draw[DynOutline,line width=0.42pt] (458.0000,182.0000) -- (460.0000,177.0000) -- (463.0000,173.0000) -- (468.0000,171.0000) -- (473.0000,171.0000) -- (478.0000,175.0000) -- (480.0000,180.0000);
\draw[DynOutline,line width=0.42pt] (438.0000,245.0000) -- (443.0000,242.0000) -- (451.0000,239.0000) -- (457.0000,240.0000) -- (464.0000,241.0000) -- (469.0000,243.0000);
\draw[DynOutline,line width=0.42pt] (437.0000,251.0000) -- (442.0000,250.0000) -- (448.0000,250.0000) -- (456.0000,250.0000) -- (463.0000,251.0000) -- (469.0000,253.0000);
\draw[DynOutline,line width=0.42pt] (433.0000,280.0000) -- (441.0000,279.0000) -- (446.0000,278.0000) -- (453.0000,279.0000) -- (459.0000,280.0000) -- (465.0000,283.0000);
\draw[DynOutline,line width=0.42pt] (433.0000,290.0000) -- (438.0000,290.0000) -- (445.0000,290.0000) -- (451.0000,290.0000) -- (457.0000,290.0000) -- (463.0000,292.0000);
\draw[DynOutline,line width=0.42pt] (431.0000,320.0000) -- (436.0000,317.0000) -- (443.0000,316.0000) -- (449.0000,316.0000) -- (456.0000,317.0000) -- (461.0000,319.0000);
\draw[DynOutline,line width=0.42pt] (431.0000,331.0000) -- (437.0000,329.0000) -- (443.0000,327.0000) -- (450.0000,327.0000) -- (456.0000,327.0000) -- (461.0000,328.0000);
\draw[DynOutline,line width=0.42pt] (431.0000,355.0000) -- (436.0000,353.0000) -- (443.0000,352.0000) -- (450.0000,352.0000) -- (456.0000,352.0000) -- (461.0000,354.0000);
\draw[DynOutline,line width=0.42pt] (432.0000,365.0000) -- (438.0000,363.0000) -- (445.0000,362.0000) -- (453.0000,362.0000) -- (458.0000,362.0000) -- (463.0000,364.0000);
\draw[DynOutline,line width=0.42pt] (272.3000,154.7000) -- (269.8250,163.7000) -- (272.3000,173.7000) -- (274.8250,182.7000) -- (273.3500,190.7000) -- (270.3500,196.7000) -- (264.8250,205.7000) -- (261.3000,213.7000) -- (260.3500,222.7000) -- (265.8250,225.7000);
\draw[DynOutline,line width=0.42pt] (272.3000,154.7000) -- (278.3500,163.7000) -- (284.3000,172.7000) -- (287.8250,182.7000) -- (287.8250,190.7000) -- (288.3500,200.7000) -- (286.3500,209.7500) -- (281.3500,217.7000) -- (273.8250,223.7000) -- (265.3500,225.7000);
\draw[DynOutline,line width=0.42pt] (250.3500,188.6500) -- (244.3500,190.7000) -- (236.3500,195.7000) -- (229.8750,201.7500) -- (226.8250,211.7000) -- (225.4000,219.7500) -- (226.8250,228.7000) -- (230.8750,234.7500) -- (236.8250,240.6500) -- (243.8750,243.6500) -- (252.3500,246.6000) -- (260.3000,250.5000) -- (267.9250,257.2500) -- (276.0750,263.0000) -- (286.0250,270.7000) -- (295.1250,276.4500) -- (303.3250,282.2500) -- (311.6500,290.0500) -- (318.4000,300.1000) -- (322.7250,311.2000) -- (324.7750,321.9500) -- (327.1250,330.2000) -- (332.4500,340.2000) -- (336.8250,348.4500) -- (340.6750,360.4500) -- (342.1000,379.1000);
\draw[DynOutline,line width=0.42pt] (404.0500,380.0000) -- (402.1500,348.9000) -- (354.5000,355.9000) -- (359.6500,380.0000);
\draw[DynOutline,line width=0.42pt] (438.0500,122.9500) -- (423.8500,309.9000) -- (388.5000,306.9500) -- (381.4500,327.5500) -- (383.3000,345.8000) -- (354.4500,348.9000) -- (357.4000,328.0000) -- (379.6000,305.0500);
\draw[DynOutline,line width=0.42pt] (250.0000,94.0000) -- (244.2500,101.1500) -- (236.5000,106.3000) -- (228.7500,113.4000) -- (224.0000,125.1500) -- (226.1500,134.0500) -- (232.1000,140.8500) -- (241.0000,144.6500) -- (248.8500,145.5000) -- (255.7000,143.3000) -- (260.6500,138.3000) -- (268.4500,130.4500) -- (273.4500,123.5500) -- (280.3000,115.6000) -- (287.1000,107.7500) -- (292.0000,97.0000);
\draw[DynOutline,line width=0.42pt] (250.3500,188.6500) -- (251.4500,193.2500) -- (246.5000,189.2500) -- (235.8500,184.2500) -- (228.1000,179.1500) -- (222.3000,170.3000) -- (222.3500,162.4000) -- (225.4000,153.6000) -- (232.3500,148.6000) -- (244.0500,149.4500) -- (253.9000,152.4500) -- (262.6500,158.5500) -- (266.5500,169.4500) -- (270.4000,184.2500) -- (271.5000,198.0500) -- (267.5000,208.0500);
\draw[DynOutline,line width=0.42pt] (325.0000,113.0000) -- (333.0000,113.0000) -- (344.0000,116.0000) -- (353.9450,120.9450) -- (363.0450,125.9450) -- (373.2450,132.0450) -- (382.3450,136.8450) -- (392.6350,139.7100) -- (402.9700,141.3200) -- (410.0700,142.7300) -- (415.2800,145.3950) -- (417.1250,153.4500) -- (417.0250,161.3500) -- (416.8250,169.2500) -- (418.8900,178.3300) -- (425.0900,181.1650) -- (429.9350,179.0100) -- (433.9800,174.1650) -- (434.9350,167.1200) -- (436.0350,158.0400) -- (437.1350,151.0400) -- (437.2450,144.9850) -- (438.3900,135.9950) -- (438.3800,128.8500) -- (434.2250,125.0950) -- (427.0800,123.3300) -- (416.5900,123.8200) -- (406.1000,124.2100) -- (398.0550,124.5450) -- (388.9450,121.5450) -- (379.9450,121.6450) -- (372.2000,121.8450) -- (364.0000,117.9000) -- (357.1000,115.0000);
\draw[DynOutline,line width=0.42pt] (379.6000,305.0500) -- (379.8000,295.2000) -- (380.8000,286.2500) -- (381.9500,275.5500) -- (386.0500,266.6500) -- (391.1000,257.9500) -- (391.3500,248.1500) -- (387.6000,245.1000) -- (383.5500,247.9500) -- (381.2000,255.8000) -- (377.0500,264.5500) -- (373.7500,273.3000) -- (366.4500,280.1000) -- (361.6000,257.2000) -- (368.8000,255.2000) -- (375.1000,250.2500) -- (379.4000,244.2500) -- (380.0000,235.3500) -- (384.8500,227.3000) -- (392.3500,219.1500) -- (399.8000,212.0500) -- (404.4500,203.0000) -- (408.0500,192.9500) -- (409.7500,182.0000) -- (407.6000,169.2000) -- (403.4000,158.4000) -- (395.1500,149.8000) -- (384.9500,145.1500) -- (375.8000,140.5500) -- (362.8000,138.8000) -- (348.9000,138.9000) -- (336.9500,142.8500) -- (327.8500,147.7000) -- (319.8000,152.6000) -- (313.7000,159.4500) -- (312.6000,167.3000) -- (315.5000,175.1500) -- (318.4500,183.0000) -- (315.8000,188.0500) -- (287.8250,182.7000);
\draw[DynThin,line width=0.40pt] (350.0000,200.0000) -- (432.1976,200.0000);
\draw[DynThin,line width=0.40pt] (350.0000,200.0000) -- (425.8081,184.9208);
\draw[DynThin,line width=0.40pt] (350.0000,200.0000) -- (419.9247,171.0362);
\draw[DynThin,line width=0.40pt] (350.0000,200.0000) -- (414.0603,157.1963);
\draw[DynThin,line width=0.40pt] (350.0000,200.0000) -- (407.7339,142.2661);
\draw[DynThin,line width=0.40pt] (350.0000,200.0000) -- (390.6532,139.1582);
\draw[DynThin,line width=0.40pt] (350.0000,200.0000) -- (377.2687,134.1674);
\draw[DynThin,line width=0.40pt] (350.0000,200.0000) -- (364.5513,126.8458);
\draw[DynThin,line width=0.40pt] (350.0000,200.0000) -- (350.0000,118.9834);
\draw[DynThin,line width=0.40pt] (350.0000,200.0000) -- (336.3659,113.9180);
\draw[DynThin,line width=0.40pt] (350.0000,200.0000) -- (322.1887,114.4056);
\draw[DynThin,line width=0.40pt] (350.0000,200.0000) -- (309.6847,120.8768);
\draw[DynThin,line width=0.40pt] (350.0000,200.0000) -- (298.9325,129.7117);
\draw[DynThin,line width=0.40pt] (350.0000,200.0000) -- (291.4383,134.5370);
\draw[DynThin,line width=0.40pt] (275.0000,125.0000) -- (317.9004,184.0473);
\draw[DynThin,line width=0.40pt] (275.0000,125.0000) -- (306.1891,186.2120);
\draw[DynThin,line width=0.40pt] (275.0000,125.0000) -- (294.1403,183.9077);
\draw[DynThin,line width=0.40pt] (275.0000,125.0000) -- (282.0052,169.2288);
\draw[DynThin,line width=0.40pt] (275.0000,125.0000) -- (275.0000,158.7165);
\draw[DynGrid,line width=0.65pt] (368.6814,316.3375) -- (385.2873,316.3375);
\draw[DynGrid,line width=0.65pt] (358.6817,326.6750) -- (381.7495,326.6750);
\draw[DynGrid,line width=0.65pt] (356.1279,337.0125) -- (382.4092,337.0125);
\draw[DynGrid,line width=0.65pt] (354.4500,348.9000) -- (383.3000,345.8000);
\draw[DynGrid,line width=0.65pt] (399.9399,211.7778) -- (431.3030,211.7778);
\draw[DynGrid,line width=0.65pt] (388.2958,223.5556) -- (430.4084,223.5556);
\draw[DynGrid,line width=0.65pt] (386.0911,235.3333) -- (429.5138,235.3333);
\draw[DynGrid,line width=0.65pt] (387.2869,247.1111) -- (428.6192,247.1111);
\draw[DynGrid,line width=0.65pt] (385.4532,258.8889) -- (427.7246,258.8889);
\draw[DynGrid,line width=0.65pt] (383.6196,270.6667) -- (426.8300,270.6667);
\draw[DynGrid,line width=0.65pt] (381.7859,282.4444) -- (425.9354,282.4444);
\draw[DynGrid,line width=0.65pt] (379.9522,294.2222) -- (425.0408,294.2222);
\draw[DynGrid,line width=0.65pt] (379.6000,305.0500) -- (388.5000,306.9500);
\draw[DynGrid,line width=0.65pt] (264.8000,164.5590) -- (264.8000,134.1234);
\draw[DynGrid,line width=0.65pt] (254.6000,152.9380) -- (254.6000,143.6533);
\draw[DynGrid,line width=0.65pt] (244.4000,149.5566) -- (244.4000,145.0182);
\draw[DynGrid,line width=0.65pt] (234.2000,148.7344) -- (234.2000,141.7466);
\draw[DynGrid,line width=0.65pt] (224.0000,157.6393) -- (224.0000,125.1500);
\draw[DynGrid,line width=0.65pt] (405.5246,200.0000) -- (432.1976,200.0000);
\draw[DynGrid,line width=0.65pt] (408.7704,188.3098) -- (425.8081,184.9208);
\draw[DynGrid,line width=0.65pt] (408.6901,175.6898) -- (419.9247,171.0362);
\draw[DynGrid,line width=0.65pt] (405.2272,163.0984) -- (414.0603,157.1963);
\draw[DynGrid,line width=0.65pt] (397.6226,152.3774) -- (407.7339,142.2661);
\draw[DynGrid,line width=0.65pt] (386.2528,145.7439) -- (390.6532,139.1582);
\draw[DynGrid,line width=0.65pt] (374.6871,140.4002) -- (377.2687,134.1674);
\draw[DynGrid,line width=0.65pt] (362.1725,138.8045) -- (364.5513,126.8458);
\draw[DynGrid,line width=0.65pt] (350.0000,138.8921) -- (350.0000,118.9834);
\draw[DynGrid,line width=0.65pt] (340.7494,141.5941) -- (336.3659,113.9180);
\draw[DynGrid,line width=0.65pt] (332.2455,145.3573) -- (322.1887,114.4056);
\draw[DynGrid,line width=0.65pt] (324.4167,149.7899) -- (309.6847,120.8768);
\draw[DynGrid,line width=0.65pt] (317.4661,155.2209) -- (298.9325,129.7117);
\draw[DynGrid,line width=0.65pt] (313.0272,164.2515) -- (291.4383,134.5370);
\draw[DynGrid,line width=0.65pt] (317.9004,184.0473) -- (282.9920,136.0000);
\draw[DynGrid,line width=0.65pt] (306.1891,186.2120) -- (280.6048,136.0000);
\draw[DynGrid,line width=0.65pt] (294.1403,183.9077) -- (278.5741,136.0000);
\draw[DynGrid,line width=0.65pt] (282.0052,169.2288) -- (276.7775,136.2225);
\draw[DynGrid,line width=0.65pt] (275.0000,158.7165) -- (275.0000,138.0000);
\fill[DynOutline] (350.0000,200.0000) circle[radius=1.65];
\node[font=\scriptsize,anchor=west] at (358.0000,207.0000) {P1};
\fill[DynOutline] (275.0000,125.0000) circle[radius=1.65];
\node[font=\scriptsize,anchor=west] at (283.0000,132.0000) {P2};
\fill[DynOutline] (368.8750,347.3500) circle[radius=1.65];
\node[font=\scriptsize,anchor=west] at (376.8750,354.3500) {G};
\fill[DynOutline] (383.3000,345.8000) circle[radius=1.65];
\fill[DynOutline] (354.4500,348.9000) circle[radius=1.65];
\fill[DynOutline] (224.0000,125.1500) circle[radius=1.65];
\fill[DynOutline] (222.3500,162.4000) circle[radius=1.65];
\end{tikzpicture}}
\caption{/u/}
\end{subfigure}
\caption{The model's sampling grid for the vowel targets /i/ and /u/.
P1 and P2 are the fixed posterior and anterior fan origins; pale lines show
construction rays and darker blue segments the sampled cavity distances.
G is the midpoint of the two inferior glottal reference points. The labial
reference points delimit the adaptive lip region. The same regional line
counts are used for both configurations; their intersections with the
articulator contours change. The diagrams are regenerated from the current
model geometry, rather than traced from screenshots.}
\label{fig:grid-system}
\end{figure}
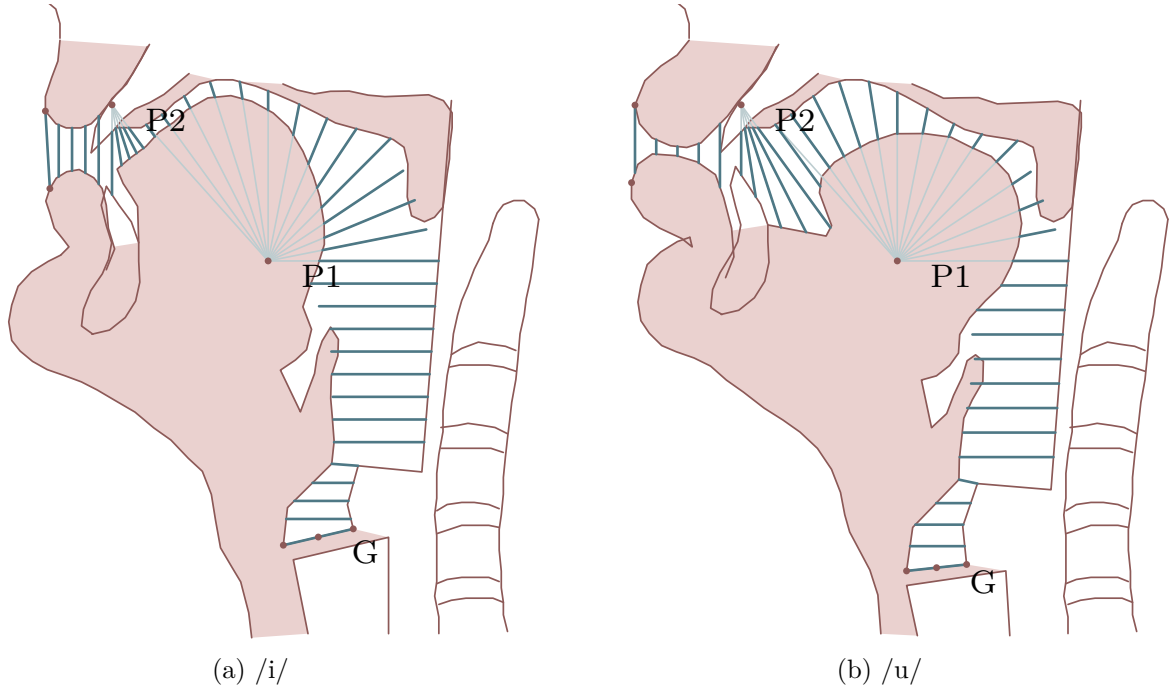

The midsagittal contours are sampled by a regionally organized grid, illustrated for /i/ and /u/ in Fig.~\ref{fig:grid-system}. The construction uses two fixed fan origins in model coordinates. The posterior origin (P1) defines rays sampling the velar and palatal regions; the anterior origin (P2), near the upper incisor, defines the fan in the anterior oral cavity. The rays intersect the relevant inner and outer contour polylines. Consequently, a fixed sampling direction does not imply a fixed cavity distance: the intersections move as the tongue, velum, jaw, and lips change configuration.

Below the posterior fan, parallel grid lines sample the pharyngeal and laryngeal regions. Their vertical spacing adapts to reference points on the current laryngeal contour. The midpoint of the two inferior reference points identifies the glottal end (G); the final measurement itself joins these two points and can therefore be slightly oblique. A further pair defines the transition between the pharyngeal and laryngeal sampling blocks. At the lips, five parallel vertical lines are distributed uniformly over a labial interval whose extent follows the outer upper-lip position relative to the fixed anterior reference. Their intersections with the upper and lower contours determine the labial cavity distances. Thus the construction combines fixed fan origins with adaptive contour intersections and adaptive parallel sampling, rather than imposing a globally equidistant grid on the entire tract.

The implementation retains the same regional sampling counts across configurations: four laryngeal lines, nine pharyngeal lines, eight velar lines, six palatal lines including a transition line, five anterior/alveolar lines, and five labial lines. This gives 37 ordered samples from the inferior tract to the lips. For a grid line with inner and outer intersections $\mathbf{q}^{\mathrm{in}}_j$ and $\mathbf{q}^{\mathrm{out}}_j$, the cavity distance and its midpoint are
\begin{equation}
 d_j=\lVert\mathbf{q}^{\mathrm{out}}_j-\mathbf{q}^{\mathrm{in}}_j\rVert,
 \qquad
 \mathbf{m}_j=\tfrac12(\mathbf{q}^{\mathrm{out}}_j+\mathbf{q}^{\mathrm{in}}_j).
\end{equation}
Contour-intersection rules additionally identify closure and set the corresponding distance to zero. A selectively smoothed sequence of midpoints defines a centerline. Longitudinal position is accumulated along this centerline and converted to the model's physical length scale. The distance samples themselves remain in contour-coordinate units; they are not direct measurements in millimeters.

Cross-sectional area is estimated from these distances using the model-specific mapping
\begin{align}
 A_j^{(0)}&=\alpha d_j^{\beta},\qquad \alpha=0.5,\quad\beta=1.5,\\
 A_j&=A_j^{(0)}\left[1+\frac{1}{1+\exp\{-(A_j^{(0)}-80)/15\}}\right].
 \label{eq:distance-area}
\end{align}
The numerical constants refer to the calibrated area scale in $\mathrm{mm^2}$. The second expression increases larger cavity areas smoothly toward twice the initial estimate while only weakly affecting narrow sections. This is an empirical distance-to-area conversion, not an independently measured three-dimensional reconstruction. The coefficient $\alpha$ includes the conversion from the internal distance units to the chosen area scale.

\begin{figure}[p]
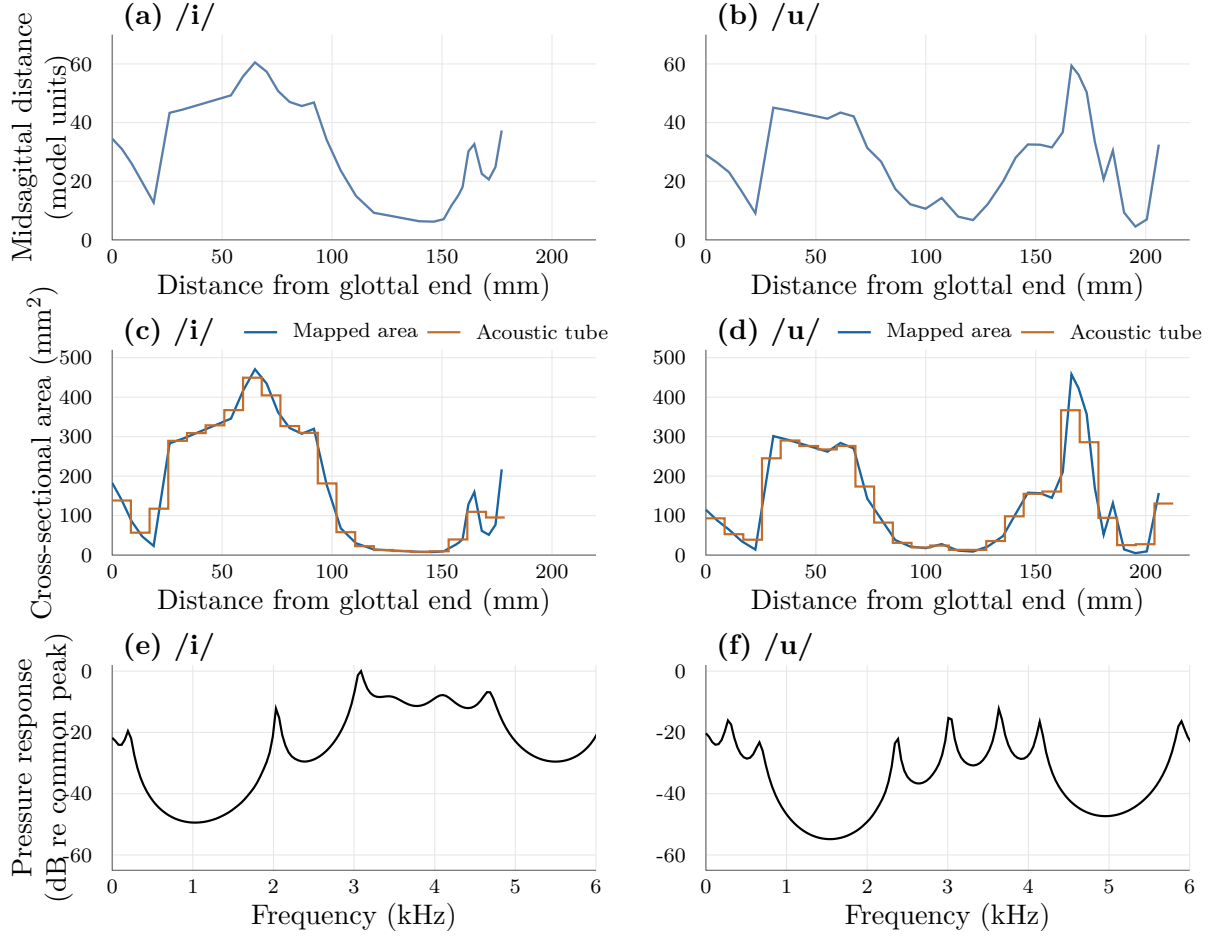

\centering
\resizebox{\linewidth}{!}{
}
\caption{Distance-to-area conversion and acoustic consequences for the same
/i/ (left) and /u/ (right) model targets as in Fig.~\ref{fig:grid-system}.
(a,b) Sampled midsagittal distances, expressed in the original contour units.
(c,d) Mapped areas (blue) and fixed-length acoustic tube sections (orange).
(e,f) Tract-inlet flow-impulse to internal lip-pressure response magnitudes,
computed with the model's fixed IR glottal-boundary convention and displayed
relative to one common maximum over both curves in the plotted frequency band.
They are not phonatory spectra or far-field radiated transfer functions.
The velopharyngeal port is closed in both configurations; no nasal branch,
friction arrow, or application annotation is displayed.}
\label{fig:distance-area-transfer}
\end{figure}

The resulting area function is resampled into fixed-length acoustic tube sections. Segment averages are calculated from multiple positions within each section, while complete closures detected in the original geometry are explicitly preserved so that spatial averaging cannot turn a narrow closed region into an open tube. \hyperref[fig:distance-area-transfer]{Figure~\ref*{fig:distance-area-transfer}} separates three quantities that are overlaid in the application: the midsagittal distance function, the mapped area function, and its piecewise-constant acoustic approximation. It also shows the associated acoustic pressure-response magnitude. The latter is calculated from an inlet flow impulse and a pressure probe at the lip end of the tube; it is a separate acoustic calculation, not another distance or area curve. Both vowel examples use a closed velopharyngeal port, so the nasal tract is omitted from these illustrations.

The acoustic discretization uses sections of length $\Delta x=8.5\,\mathrm{mm}$ and an internal sampling rate of $40\,\mathrm{kHz}$. This gives a nominal propagation speed $\Delta x f_s=340\,\mathrm{m\,s^{-1}}$. The current impedance and aerodynamic-compliance calculations use $350\,\mathrm{m\,s^{-1}}$ as their sound-speed constant. These values are implementation conventions, not a temperature-calibrated material model. Section area $A_j$ determines characteristic impedance through
\begin{equation}
 Z_j=\frac{\rho c_Z}{A_j},
 \label{eq:impedance}
\end{equation}
where areas are expressed in SI units, $\rho=1.14\,\mathrm{kg\,m^{-3}}$, and $c_Z=350\,\mathrm{m\,s^{-1}}$. The solver stores traveling volume-flow waves; acoustic pressure and net flow are recovered from their sum and difference. At an ordinary two-section junction, the corresponding reflection coefficient is
\begin{equation}
 r_j=\frac{Z_{j+1}-Z_j}{Z_{j+1}+Z_j}.
\end{equation}
Propagation includes adjustable attenuation and frequency-dependent smoothing. Narrow oral sections also receive an empirical transmission loss, distinct from the mean-flow resistance described below. Boundary reflection and radiation are parameterized separately.

\begin{figure}[p]
\centering
\resizebox{\linewidth}{!}{\begin{tikzpicture}[x=0.052cm,y=0.011cm,font=\small]
\draw[DynArea,line width=.7pt] (0.000,128.439) -- (4.785,100.808) -- (9.754,67.438) -- (15.099,35.994) -- (20.998,15.429) -- (28.415,238.377) -- (34.740,225.822) -- (40.215,212.879) -- (45.690,199.511) -- (51.132,185.729) -- (56.559,215.791) -- (61.992,240.871) -- (67.454,207.565) -- (73.021,189.480) -- (78.727,70.792) -- (84.773,35.463) -- (90.481,19.579) -- (96.349,11.984) -- (102.481,11.079) -- (108.722,18.580) -- (115.174,34.424) -- (121.656,65.424) -- (127.692,116.437) -- (133.299,149.159) -- (138.973,130.777) -- (144.306,96.668) -- (149.023,90.573) -- (152.576,129.327) -- (156.053,158.396) -- (158.326,410.118) -- (160.498,207.310) -- (165.118,65.294) -- (169.648,19.971) -- (173.981,0.000) -- (178.282,0.000) -- (182.589,0.000) -- (186.260,28.609);
\draw[DynTube,line width=.8pt] (0.000,0.000) -- (0.000,103.128) -- (8.500,103.128) -- (8.500,50.421) -- (17.000,50.421) -- (17.000,54.778) -- (25.500,54.778) -- (25.500,219.619) -- (34.000,219.619) -- (34.000,217.488) -- (42.500,217.488) -- (42.500,196.772) -- (51.000,196.772) -- (51.000,208.112) -- (59.500,208.112) -- (59.500,226.224) -- (68.000,226.224) -- (68.000,179.446) -- (76.500,179.446) -- (76.500,63.419) -- (85.000,63.419) -- (85.000,25.000) -- (93.500,25.000) -- (93.500,25.000) -- (102.000,25.000) -- (102.000,25.000) -- (110.500,25.000) -- (110.500,35.403) -- (119.000,35.403) -- (119.000,80.440) -- (127.500,80.440) -- (127.500,136.163) -- (136.000,136.163) -- (136.000,120.996) -- (144.500,120.996) -- (144.500,102.315) -- (153.000,102.315) -- (153.000,229.431) -- (161.500,229.431) -- (161.500,75.071) -- (170.000,75.071) -- (170.000,0.000) -- (178.500,0.000) -- (178.500,0.000) -- (187.000,0.000);
\draw[teal!80!black,line width=.8pt] (93.500,400.000) -- (93.500,480.000) -- (102.000,480.000) -- (102.000,465.000) -- (110.500,465.000) -- (110.500,485.000) -- (119.000,485.000) -- (119.000,515.000) -- (127.500,515.000) -- (127.500,545.000) -- (136.000,545.000) -- (136.000,565.000) -- (144.500,565.000) -- (144.500,550.000) -- (153.000,550.000) -- (153.000,525.000) -- (161.500,525.000) -- (161.500,505.000) -- (170.000,505.000) -- (170.000,485.000) -- (178.500,485.000) -- (178.500,468.000) -- (187.000,468.000) -- (187.000,454.000) -- (195.500,454.000) -- (195.500,442.000) -- (204.000,442.000) -- (204.000,434.000) -- (212.500,434.000);
\draw[black!50,->] (0.000,0.000) -- (225.000,0.000);
\draw[black!50,->] (0.000,0.000) -- (0.000,650.000);
\draw[teal!50!black] (93.500,400.000) -- (212.500,400.000);
\draw[violet,->,line width=1.2pt] (93.500,33.000) -- (93.500,400.000);
\draw[violet,line width=1.8pt] (93.500,480.000) -- (102.000,480.000);
\node[anchor=east,text=violet] at (89.500,216.500) {\shortstack[r]{Velopharyngeal\\junction}};
\node[anchor=west,text=teal!80!black] at (93.500,603.000) {Equivalent nasal branch (both passages)};
\node[anchor=west,text=violet,font=\scriptsize] at (105.000,500.000) {Port: 80 mm$^2$};
\node[anchor=west,font=\scriptsize] at (3.000,28.000) {Glottis};
\node[anchor=east,font=\scriptsize] at (184.000,28.000) {Lips};
\node[anchor=east,font=\scriptsize,text=teal!80!black] at (212.500,372.000) {Nostrils};
\node[anchor=east,font=\scriptsize] at (-3.000,0.000) {0};
\draw[black!50] (-1.000,0.000) -- (0.000,0.000);
\node[anchor=east,font=\scriptsize] at (-3.000,100.000) {100};
\draw[black!50] (-1.000,100.000) -- (0.000,100.000);
\node[anchor=east,font=\scriptsize] at (-3.000,200.000) {200};
\draw[black!50] (-1.000,200.000) -- (0.000,200.000);
\node[anchor=east,font=\scriptsize] at (-3.000,300.000) {300};
\draw[black!50] (-1.000,300.000) -- (0.000,300.000);
\node[anchor=east,font=\scriptsize,text=teal!80!black] at (90.500,400.000) {0};
\node[anchor=east,font=\scriptsize,text=teal!80!black] at (90.500,500.000) {100};
\node[anchor=east,font=\scriptsize,text=teal!80!black] at (90.500,600.000) {200};
\node[font=\scriptsize] at (0.000,-23.000) {0};
\draw[black!50] (0.000,-5.000) -- (0.000,0.000);
\node[font=\scriptsize] at (25.000,-23.000) {25};
\draw[black!50] (25.000,-5.000) -- (25.000,0.000);
\node[font=\scriptsize] at (50.000,-23.000) {50};
\draw[black!50] (50.000,-5.000) -- (50.000,0.000);
\node[font=\scriptsize] at (75.000,-23.000) {75};
\draw[black!50] (75.000,-5.000) -- (75.000,0.000);
\node[font=\scriptsize] at (100.000,-23.000) {100};
\draw[black!50] (100.000,-5.000) -- (100.000,0.000);
\node[font=\scriptsize] at (125.000,-23.000) {125};
\draw[black!50] (125.000,-5.000) -- (125.000,0.000);
\node[font=\scriptsize] at (150.000,-23.000) {150};
\draw[black!50] (150.000,-5.000) -- (150.000,0.000);
\node[font=\scriptsize] at (175.000,-23.000) {175};
\draw[black!50] (175.000,-5.000) -- (175.000,0.000);
\node[font=\scriptsize] at (200.000,-23.000) {200};
\draw[black!50] (200.000,-5.000) -- (200.000,0.000);
\node[font=\scriptsize] at (225.000,-23.000) {225};
\draw[black!50] (225.000,-5.000) -- (225.000,0.000);
\node[] at (112.500,-65.000) {Distance from glottis along oral path (mm)};
\node[rotate=90] at (-15.000,325.000) {Cross-sectional area (mm$^2$)};
\node[anchor=west,text=DynArea,font=\scriptsize] at (5.000,632.000) {Mapped oral area};
\node[anchor=west,text=DynTube,font=\scriptsize] at (5.000,605.000) {Oral acoustic tube};
\end{tikzpicture}}
\caption{Area functions for the bilabial nasal /m/ with the velum lowered to open the nasal port. The oral path is closed at the lips while the nasal port remains open. The blue curve is the mapped oral area function; the orange steps are the effective acoustic tube sections. The green steps show the equivalent nasal branch, including its coupling section in purple. The branch origin is aligned horizontally with the actual junction in the oral--pharyngeal tube, indicated by the arrow. Both profiles use the same length and area scales; the nasal baseline is shifted upward for display, with its own area labels. Horizontal distance along the nasal profile is measured from the junction and does not denote an anatomical continuation of the oral axis. The arrow marks connectivity, not a one-way flow constraint.}
\label{fig:nasal-area-m}
\end{figure}
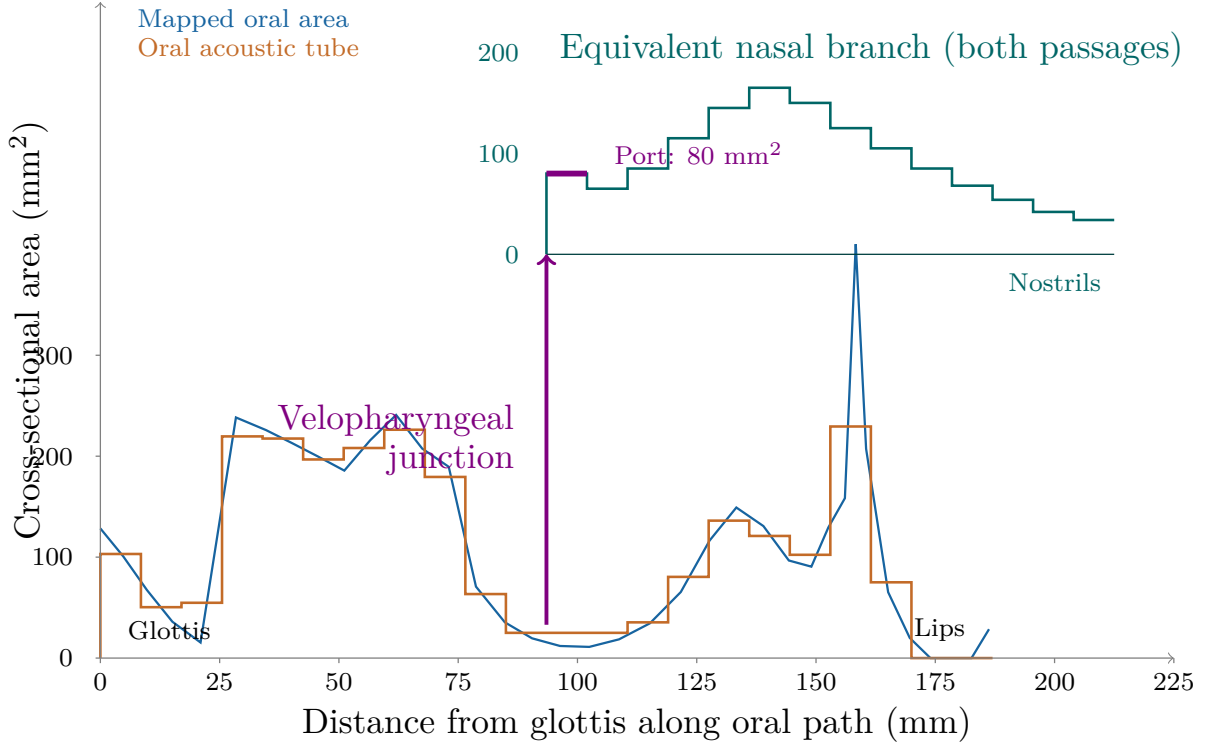

The nasal tract is represented by one equivalent branch for both nasal passages. Its junction is located from the tract geometry, and the velopharyngeal port area varies smoothly with velum control. The remaining branch areas are fixed in the current implementation. A raised velum disconnects this branch; lowering the velum opens the port up to the present reference area of $80\,\mathrm{mm^2}$. At the three-way acoustic junction, a common pressure is obtained from the incident waves and branch admittances. Scattered waves are returned to both oral directions and to the nasal branch. The nose is therefore part of the coupled resonator, rather than an independently mixed nasal filter. \hyperref[fig:nasal-area-m]{Figure~\ref*{fig:nasal-area-m}} shows the oral and nasal area functions for the bilabial nasal /m/ with lip closure and a lowered velum, with the nasal branch positioned directly above its coupling junction.

Additional corrections represent pathways that are not fully visible in the midsagittal contour. For laterals, an effective minimum area models the combined left and right passages; its default is $25\,\mathrm{mm^2}$. A separate, spatially restricted bypass correction prevents a lowered velum from causing an unintended sagittal blockage when no deliberate dorsal consonantal constriction is present. These corrections change the effective tube geometry and consequently influence both sound propagation and flow. They do not resolve the individual lateral channels as separate acoustic branches.

Further empirical corrections concern tooth obstacles and narrow fricative grooves. Their areas and locations can be adjusted in the application. In the continuous oral-friction mode documented here, however, a plosive-release marker does not insert an artificial fixed-area burst constriction. The acoustic release geometry is taken from the evolving configuration. The visual closure and release markers remain available independently. Ordinary oral velum support yields to obstruent raising and nasal lowering: its activation is reduced during the onset ramp of a priority gesture, remains suppressed during its full activation, and is released during its offset ramp if the oral gesture is still active. The same coordinated envelopes are used for trajectory generation and gesture display.

\section{Aerodynamic Model}
\subsection{Pressure, flow, and boundary conditions}
The aerodynamic component is a lumped, quasi-steady flow network with a dynamic pressure-storage state. Ambient pressure is the reference at the oral and nasal outlets. The pulmonary control specifies a subglottal pressure $P_s$; its present mapping covers approximately $0$ to $981\,\mathrm{Pa}$. Glottal area $A_g(t)$ is prescribed by the excitation model. A forward-flow orifice relation provides the glottal inflow,
\begin{equation}
 U_g=C_d A_g\sqrt{\frac{2[P_s-P_{\mathrm{in}}]_+}{\rho}},
 \qquad [x]_+=\max(x,0),
 \label{eq:glottflow}
\end{equation}
with discharge coefficient $C_d=0.8$. Reverse flow is not represented by this relation. The effective inlet pressure $P_{\mathrm{in}}$ used by the acoustic source combines the storage-pressure estimate with a smoothed estimate of acoustic inlet loading, as specified in Section~\ref{sec:coupling}.

Each tract flow path includes linear viscous resistance and a quadratic pressure loss. For circular equivalent sections, the linear contribution is
\begin{equation}
 R=\sum_j\frac{8\pi\mu\ell_j}{A_j^2},
 \qquad \mu=1.86\times10^{-5}\,\mathrm{Pa\,s}.
 \label{eq:resistance}
\end{equation}
The quadratic term is assigned once per flow path using its narrowest section,
\begin{equation}
 \Delta P=RU+BU^2,\qquad
 B=\frac{\rho}{2C_d^2 A_{\min}^2}.
 \label{eq:quadratic}
\end{equation}
This avoids counting one extended constriction as multiple independent jets merely because it spans several tube sections. These expressions provide equivalent resistive loads; they are not a resolved model of flow separation or turbulent dissipation in an anatomically irregular channel. A completely closed path has infinite resistance and zero outflow.

With an open velopharyngeal port, the shared pharyngeal path is in series with parallel oral and nasal paths. The junction pressure is found numerically so that branch flow and the pressure drop along the shared path are consistent. Glottal inflow thus need not vanish when the oral outlet is closed: air may leave through the nose. Conversely, simultaneous oral and nasal closure permits supraglottal pressure to build up. These distinctions are essential for separating nasals from oral stops.

Air connected to the glottis supplies a lumped compliance
\begin{equation}
 C=\frac{V_{\mathrm{conn}}}{\rho c_A^2},
 \qquad c_A=350\,\mathrm{m\,s^{-1}},
\end{equation}
where $V_{\mathrm{conn}}$ includes the accessible tract volume up to closures and any connected nasal volume. Wall compliance is not included in this storage term. The storage pressure $P$ obeys
\begin{equation}
 C\frac{dP}{dt}=U_g-U_o-U_n,
 \label{eq:storage}
\end{equation}
where $U_o$ and $U_n$ are oral and nasal outflow. The implementation advances this state implicitly, using a backward-Euler balance and a monotone pressure search. For an utterance, the pressure after one $5$-ms control frame initializes the next frame. The pressure at the end of the longer acoustic response is not used for this transfer, since doing so would advance the aerodynamic state by more than one control frame.

\subsection{Approximations and coupling to excitation}
\label{sec:coupling}
The storage pressure and the acoustic traveling waves are separate quantities. The former captures slow pressure accumulation and venting; the latter describes propagation, reflections, and resonances. During a pulse-response simulation, the glottal area varies at the internal sampling rate, and the storage network is advanced alongside the acoustic calculation. For stable source loading, the pressure entering Eq.~\eqref{eq:glottflow} is the larger of the storage pressure and a low-pass estimate of acoustic inlet pressure, limited to the interval from zero to $P_s$. These two estimates are not added as independent mean pressures. This is a numerical coupling approximation rather than an exact solution of a distributed compressible-flow system.

The tract-pressure trace used in source-amplitude estimates combines the storage pressure with an acoustic pressure probe from which a running mean has been removed. Glottal-flow and pressure traces are summarized after an initial warmup interval. Thus the scalar amplitudes driving noise are inferred from a frame-specific simulation extending beyond the frame itself; they are not instantaneous, spatially resolved measurements at the oral jet. In particular, current oral-friction drive uses glottal flow as a proxy for flow through the oral constriction. This approximation becomes less direct when flow can escape through a nasal branch.

For a held static articulation, an iterative pressure-consistency calculation can find the pressure state whose value after one control frame reproduces its initial value. For a moving articulation, the chronological transfer described above is used instead. Neither procedure integrates the mass, stiffness, or collision dynamics of vocal folds. The prescribed glottal pulse can be modified empirically by supraglottal pressure loading, but this does not make it self-oscillating. Likewise, the segmental F0 adjustment described below is imposed by control rather than inferred from aerodynamic oscillation conditions.

\begin{figure}[p]
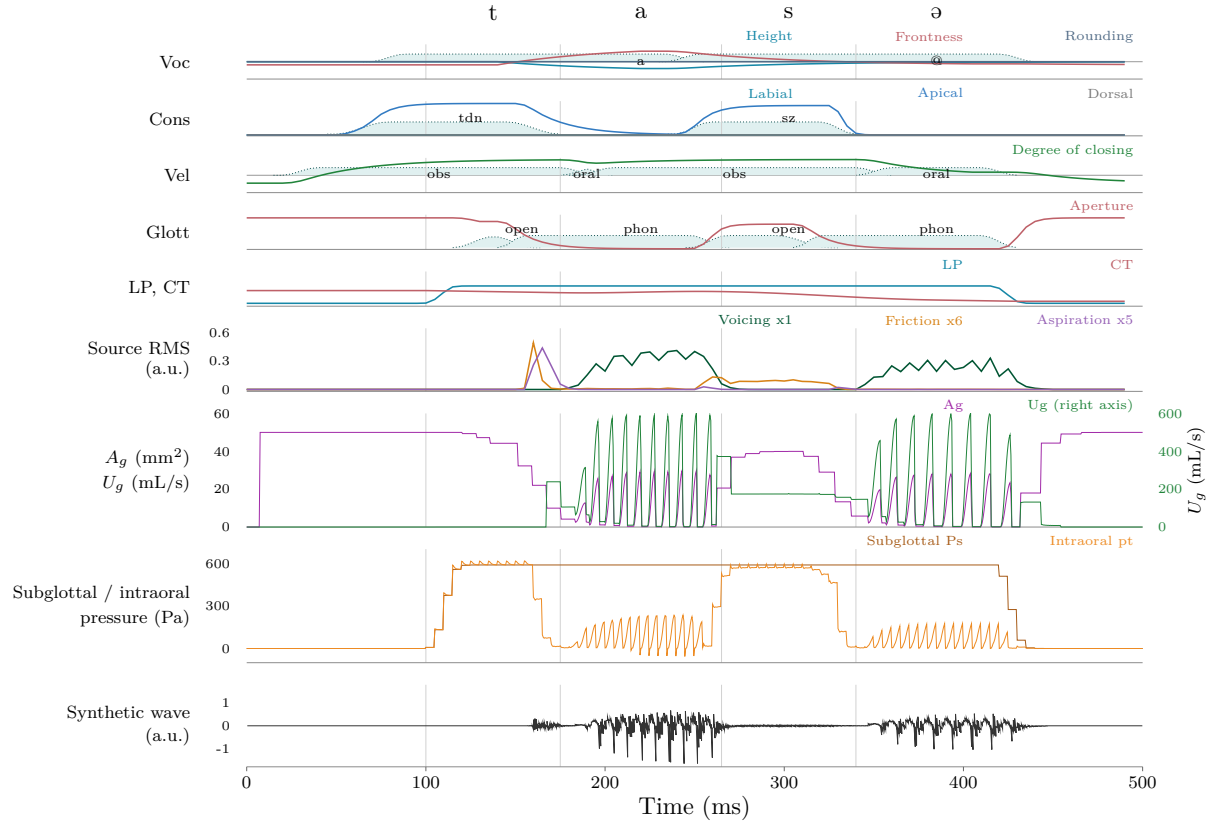

\centering
\definecolor{Exampleheight}{HTML}{1687a7}
\definecolor{Examplefront}{HTML}{be6068}
\definecolor{Exampleround}{HTML}{516b86}
\definecolor{Examplelab}{HTML}{1687a7}
\definecolor{Exampletip}{HTML}{327bc2}
\definecolor{Exampledors}{HTML}{777777}
\definecolor{Examplevel}{HTML}{21843b}
\definecolor{Examplega}{HTML}{be6068}
\definecolor{Examplelp}{HTML}{1687a7}
\definecolor{Examplect}{HTML}{be6068}
\definecolor{Exampleasp}{HTML}{9560b5}
\definecolor{Examplephon}{HTML}{005533}
\definecolor{Exampleoral}{HTML}{d78216}
\definecolor{Exampleag}{HTML}{a12aa8}
\definecolor{Exampleug}{HTML}{21843b}
\definecolor{Exampleps}{HTML}{a75b17}
\definecolor{Examplept}{HTML}{ed8b23}
\definecolor{Examplewave}{HTML}{333333}
\resizebox{\linewidth}{!}{
}
\caption{Tasse /\textipa{\textprimstress tas@}/. The initial apical gesture reaches its target 25 ms earlier with the release time retained, placing closure before pulmonary pressure onset. The five upper tiers show vocalic, consonantal, velar, glottal, and pulmonary/CT control. Pale green envelopes with dotted outlines show the actual gesture activations, including onset and offset ramps; their heights are scaled for display. In the vocalic and velum tiers, activation zero coincides with trajectory zero. Positive velum values indicate raising/closing and negative values lowering/opening. The vertical lines mark segment-timeline reference boundaries, not gesture boundaries. The shared amplitude panel shows voicing, oral friction, and aspiration, as the absolute RMS amplitudes of their separately synthesized radiated components in non-overlapping 5-ms windows. For display only, voicing is multiplied by 1, oral friction by 6, and aspiration by 5. These fixed factors, amplitude limits, and synthesis gains are identical across all three words; no per-source or per-word peak normalization is applied. These are output amplitudes in arbitrary units, not calibrated sound pressures. Area and flow are reconstructed from responses associated with glottal pulse times; pressures use successive control frames. The final panel shows the radiated floating-point synthetic output before PCM quantization.}
\label{fig:tasse-aerodynamics}
\end{figure}

\hyperref[fig:tasse-aerodynamics]{Figures~\ref*{fig:tasse-aerodynamics}}--\ref{fig:lade-aerodynamics} illustrate the control and aerodynamic calculation for three German words with first-syllable stress: /\textipa{\textprimstress tas@}/ (``Tasse''), /\textipa{\textprimstress mu:z@}/ (``Muse''), and /\textipa{\textprimstress la:d@}/ (``Lade''). Together they illustrate voiceless and voiced plosives, voiceless and voiced fricatives, a nasal, a lateral, short and long vowels, and unstressed schwa. They sample these sound classes without constituting an exhaustive inventory of German phonemes.

Each overview combines the gestural control tiers, a shared source-amplitude panel, aerodynamic diagnostics, and the radiated signal on one time axis. The source panel shows absolute RMS amplitudes of the separately synthesized voicing, oral-friction, and aspiration components in non-overlapping $5$-ms windows. For visibility, fixed display multipliers of $1$ for voicing, $6$ for oral friction, and $5$ for aspiration are applied. These factors and the amplitude-axis limits are identical across all three words; synthesis gains and audio are unchanged. No component or utterance is normalized to its own peak. These radiated-output amplitudes share arbitrary output units; they are not calibrated sound pressures. Consequently, the very small residual oral-noise component in ``Lade'' remains close to zero instead of being magnified to full scale. The pressure panel distinguishes subglottal pressure $P_s$ from intraoral pressure $p_t$.

For ``Tasse'', the initial apical closing gesture begins and reaches its target $25$ ms earlier, with its release time retained. This example-specific adjustment establishes closure before pulmonary pressure onset and removes the premature initial radiated transient.

The area and flow traces are assembled from responses associated with successive CT-derived glottal pulse times; pressure traces are assembled from consecutive control frames. They are diagnostic reconstructions of the framewise calculation, rather than traces from a single uninterrupted time-varying waveguide simulation. The curves report the implemented model, including any residual source activity outside the intended consonantal interval.

\section{Excitation and Noise Sources}
\subsection{Phonatory excitation}
Glottal excitation is constructed from a controlled area pulse with separate opening and closing portions. Raised-cosine transitions connect a rest area to a peak area and back. Open quotient, the fraction of open time assigned to closing, pulmonary pressure, and glottal-aperture control determine the nominal pulse. The application default for the closing fraction is $0.10$. Pressure loading can reduce the excursion and, if enabled, extend closing into the nominal closed interval without accelerating the opening transition. This is a phenomenological adjustment of prescribed motion.

For voiced fricatives, a dimensionless parameter $Q$ redistributes glottal excursion into a permanent opening. If $A_0$ is the unredistributed rest area and $\Delta A$ the excursion after pressure loading, the areas are
\begin{equation}
 A_{\mathrm{rest}}=A_0+Q\Delta A,\qquad
 A_{\mathrm{peak}}=A_0+\Delta A.
 \label{eq:q}
\end{equation}
The remaining excursion is $(1-Q)\Delta A$. At $Q=0$, no additional rest area is introduced. At $Q=1$, the glottis remains open at the peak area, and periodic phonatory pulse insertion is suppressed. This endpoint therefore does not represent a closed glottis. The default $Q$ is $0.3$, while the full interval $0\leq Q\leq1$ is available for listening experiments. Its application follows the drive of a voiced oral-constriction gesture group.

CT determines a nominal frequency $f_{\mathrm{CT}}$. The present calibration is linear from $80\,\mathrm{Hz}$ at CT~$500$ to $200\,\mathrm{Hz}$ at CT~$1000$, with continuation to CT~$1300$ for higher intonation. This mapping is a control convention, not a biomechanical prediction. The period timing and its use in synthesis are described in Section~\ref{sec:ola}. For each control configuration, the prescribed area pulse, subglottal pressure, inlet loading, and acoustic tube jointly produce a radiated \emph{pulse response}. Unlike a unit impulse response of a linear filter, this response already contains the modeled glottal excitation and its nonlinear flow relation.

\subsection{Oral frication and plosive release}
In the continuous mode used here, oral-noise activation depends on an open oral path and its aerodynamic drive. A fricative label or a plosive-release event is not required. The narrowest section within the designated oral part of the effective tube supplies area $A_{\min}$ for closure detection. A complete closure produces no oral source. For open geometries, the additional constriction-area gate is disabled:
\begin{equation}
 W(A_{\min})=1\qquad\text{for an open oral path}.
 \label{eq:noisearea}
\end{equation}
The resulting source-amplitude estimate is
\begin{equation}
 a_o=K_o\,\overline{|U_g|}\,\overline{[p_t]_+}.
 \label{eq:oralnoise}
\end{equation}
The bars denote averages after warmup within the current response simulation, and $p_t$ is the pressure trace described in Section~\ref{sec:coupling}. The calibration coefficient $K_o$ absorbs units and scale factors. Equation~\eqref{eq:oralnoise} is a heuristic drive law. It is not a measured turbulence-intensity relation, a Reynolds-number criterion, or a calculation of the local pressure drop at the jet.

The principal oral acoustic source is a serial pressure dipole associated with the constriction or a downstream obstacle. Its radiated kernel is computed on the current tract. Source placement between section junctions can interpolate between adjacent tube-junction responses to reduce spatial grid jumps. In the dorsal fricative configuration, an additional volume-flow monopole can contribute; its kernel is weighted by a geometric flow-to-pressure factor and empirical palatal/velar coefficients. The monopole uses an independent noise stream. These two source types are acoustic representations of excitation, not a simulation of individual turbulent vortices.

Noise spectra are shaped separately from tract filtering. Labial, apical, and dorsal source descriptions supply broad filters and empirical spectral corrections. Within the apical and dorsal regions, geometric descriptors allow gradual interpolation between anterior/posterior and palatal/velar calibrations. Some source placement and correction choices still depend on the articulatory context. The approach is therefore geometry-informed and empirically calibrated; it is not a universal source-spectrum law inferred only from fluid mechanics.

A general regional gain multiplies the complete oral source, including any associated monopole. Additional regional gains calibrate voiced frication and oral friction in voiceless-plosive contexts. The latter gain is currently $20$ for all oral regions. These context gains are assigned through gesture grouping and alter amplitude, not the aerodynamic activation condition. A scalar compensation of response-kernel gain also permits partial reduction of large differences in effective transfer gain while retaining the kernel's spectral ratios. Consequently, the final noise level should not be interpreted as an absolute prediction of acoustic sound pressure from Eq.~\eqref{eq:oralnoise} alone.

During a plosive opening transition, a source persists for as many control frames as the geometry and Eq.~\eqref{eq:oralnoise} permit. No minimum duration is imposed. Friction during closure formation is likewise not explicitly prohibited. The retained release detector marks the transition from a complete oral closure to opening for visualization; it no longer creates a fixed-area acoustic burst constriction or enables oral-noise generation. Its role in the separate post-release aspiration calibration is retained. These distinctions permit the duration of release noise to be studied independently of the visual event marker.

\subsection{Glottal aspiration and voiced-frication modulation}
Aspiration is generated at the glottal source and filtered through its own radiated tract response. Its noise stream is independent of the oral dipole stream. The scalar drive is
\begin{equation}
 a_a=K_a E_g(g)\,\overline{|U_g|}\,
              \overline{[p_s-p_t]_+},
 \label{eq:aspiration}
\end{equation}
where $g$ is glottal-aperture control on its normalized scale. The implemented aperture envelope $E_g$ uses smooth opening and closing weights within the existing glottal-friction range, approximately $0.05<g<0.30$. Its activation also respects the existing articulatory/aerodynamic eligibility conditions, including pressure and closure-related suppression. Aspiration is thus not triggered solely by a plosive-release flag. It can occur in other appropriate glottal configurations, including those used for /h/.

After an identified voiceless plosive release, an additional gain of $20$ is applied to the already present aspiration. This gain is shared across oral places of articulation. It neither inserts additional noisy frames nor changes the aperture envelope in Eq.~\eqref{eq:aspiration}. The release context ends on renewed oral closure, return to the nonaspirating glottal-control range, or the start of a later oral or glottal gesture. Ordinary /h/ and voiced-plosive contexts do not receive this additional release gain. Broad high- and low-pass source filtering provides an independently adjustable aspiration spectrum.

In voiced fricatives, continuous oral noise is amplitude-modulated relative to the glottal period. Let $\phi$ denote cycle phase, $O$ the actual open fraction of the prescribed pulse, $d$ modulation depth, $r$ the residual fraction, and $G_v$ the regional peak gain. A raised-cosine bump $b(\phi;O,\theta)$ is centered at $O/2+\theta$, with support of width $O$, interpreted cyclically. The modulation gain is
\begin{equation}
 M(\phi)=G_v\{1-d(1-r)[1-b(\phi;O,\theta)]\}.
 \label{eq:modulation}
\end{equation}
Thus $d=0$ gives an unmodulated gain $G_v$; for $d=1$, the minimum is $rG_v$ and the maximum $G_v$. The phase offset is adjustable, and the default places the maximum at the middle of the open phase. The current defaults are $d=1$ and $r=0.1$. Gesture-dependent engagement interpolates between ordinary noise and this modulation, preventing anticipatory vowel phonation from applying it to a voiceless fricative. At $Q=1$, the selected noise gain remains, but periodic modulation is suppressed. A modulated noise envelope does not require the random waveform itself to repeat from one cycle to the next.

An additional segmental control lowers frequency during voiced frication:
\begin{equation}
 f_0(t)=f_{\mathrm{CT}}(t)\,2^{-D e(t)/12},
 \label{eq:f0lowering}
\end{equation}
where $D$ is the requested reduction in semitones and $e$ a smoothed engagement signal. The default $D$ is $1.5$ semitones. Engagement follows the voiced-constriction context with time constants of $20\,\mathrm{ms}$ for increasing engagement and $100\,\mathrm{ms}$ for decreasing engagement, providing a gradual frequency recovery after the constriction. The adjustment changes the common timing used for pulse insertion and noise modulation. It is an empirical control, not an F0 change emerging from the computed pressure drop.

\section{Synthesis by Overlap-and-Add}
\label{sec:ola}
\subsection{Control frames, glottal pulse times, and acoustic response duration}
The three temporal concepts introduced above are implemented as follows. Control frames, defined by Eq.~\eqref{eq:controlgrid}, supply articulatory and aerodynamic configurations every $5\,\mathrm{ms}$. Glottal pulse times are generated by integrating the phase associated with $f_0(t)$. Acoustic response duration specifies the time span of each computed tract or pulse response and can exceed both a control frame and a glottal period. Confusing these representations would suggest, incorrectly, that the synthesis concatenates independent $5$-ms waveform fragments.

The phase calculation uses a $0.25$-ms timing step. Each completed cycle defines a glottal pulse time, which is mapped to both the internal and output sample grids. Each glottal pulse time is associated with the control frame active at that time; call this frame index $k_i$ for pulse time $i$. The CT-derived phase continues through voiceless intervals, although phonatory output is suppressed when the glottal setting provides no oscillatory component. Here, $f_0(t)$ denotes the model's commanded timing frequency, including intervals in which no acoustic F0 is present. Continuing the phase rather than restarting it at each voicing onset implements the hypothesis of an underlying CT trajectory that extends across changes in voicing. This is a control assumption, not a claim that vocal folds oscillate during voiceless segments. Continuity between observed F0 contours on either side of such an interval may motivate this interpretation, but does not by itself establish the intervening tension trajectory or a physiological pulse phase. Maintaining this timing grid does not make voiceless noise periodic.

Acoustic responses are precomputed for the control frames, rather than recalculated at each glottal pulse time. The standard acoustic response duration is $25.6\,\mathrm{ms}$, corresponding to $1024$ internal samples at $40\,\mathrm{kHz}$. Pulse simulation adds a $100$-sample ($2.5$-ms) warmup, which is discarded before pulse insertion. The response waveform is not cut to the next control-frame boundary or to the next glottal pulse time. Responses are finite, however: their simulated tails are retained only within the configured acoustic response duration and output buffer. Natural decay is not a claim of mathematically infinite support.

\subsection{Superposition of phonatory responses}
Let $p_k[m]$ be the radiated pulse response for control frame $k$, after warmup removal and conversion to the output sample rate. If $n_i$ is the output-sample index of glottal pulse time $i$, the phonatory signal is
\begin{equation}
 y_p[n]=G\sum_i a_{k_i}\,p_{k_i}[n-n_i].
 \label{eq:pulseola}
\end{equation}
Here $G$ is an overall playback calibration and $a_k$ an additional pulmonary-pressure and oscillation weighting. In the current implementation, $a_k$ is the normalized subglottal-pressure control multiplied by the aperture-dependent vibration factor. Pressure and glottal-area effects are also present inside $p_k$; the outer weighting is an explicit synthesis calibration rather than a separate physical law. The $Q=1$ endpoint excludes periodic insertion altogether, avoiding repeated startup transients from constant-area simulations.

Equation~\eqref{eq:pulseola} allows the response to one excitation to continue while later responses are added. No Hann analysis window or complementary crossfade is required to join adjacent phonatory fragments, because there are no such fragments to join. This does not exclude finite response truncation or filtering effects, which must be checked separately. Nor does linear addition of completed responses make the construction of each pulse response linear: the prescribed glottal motion, pressure storage, and inflow relation can be nonlinear within its calculation.

Each response uses a fixed oral and nasal configuration for its duration, while glottal area varies within the pulse simulation. The acoustic wave state of one frame-specific calculation is not transferred to the next. Thus an already emitted response retains the tract configuration under which it was calculated, even when a later response represents a different configuration. This is the central frozen-geometry approximation of the method. It avoids directly replacing the tube geometry around an ongoing global acoustic state, but it also omits interactions between previously excited waves and subsequent tract movement. It should not be equated with a fully time-varying waveguide or assumed to be artifact-free under arbitrary rapid articulation changes.

\subsection{Continuous noise and acoustic filtering}
Oral friction and aspiration are synthesized from continuous pseudorandom streams. They are not copied noise packets attached to individual glottal pulse times. For a single source, the construction can be written
\begin{equation}
 y_s[n]=G\sum_{\nu} x_s[\nu]\,h_{s,k(\nu)}[n-\nu],
 \label{eq:noiseola}
\end{equation}
where $h_{s,k}$ is the radiated kernel for source $s$ in control frame $k$, and $x_s$ includes colored random noise, the aerodynamic amplitude, and the applicable gain or modulation. Independent source components are calculated separately and summed. As with phonatory responses, each contribution continues beyond the control frame in which it originated.

Equation~\eqref{eq:noiseola} describes a kernel associated with the \emph{time of the individual noise-source sample}. It is not simply the repeated application of a newly selected output-time filter to the entire previous noise history. In the current implementation, each source sample launches a contribution through the kernel of its control frame. Source-filter parameters and scalar amplitude or compensation values are interpolated between adjacent control values where implemented; the full tract kernel is selected framewise rather than continuously morphed between frames. The noise random-number generators maintain continuity across these selections and use reproducible seeds.

For voiced-fricative modulation, the phase in Eq.~\eqref{eq:modulation} is calculated from the same successive glottal pulse times that define phonatory insertion. The open fraction is obtained from the corresponding pulse's open duration relative to the interval between consecutive glottal pulse times. Lengthening the period therefore slows both phonatory repetition and the oral-noise envelope. The random carrier remains stochastic. Aspiration follows its separate glottal envelope and is not subjected to this voiced-oral-frication modulation.

The internal tract simulations use $40\,\mathrm{kHz}$, whereas the standard output is $16\,\mathrm{kHz}$. Pulse responses and aspiration kernels currently use linear sample-rate conversion. Oral-noise kernels use a band-limited windowed-sinc conversion with zero extension and an explicitly retained time offset for the filter tails. This conversion window is part of the resampling filter, not a window that gates the glottal period or separates adjacent synthesis segments. The differing conversion paths are numerical features of the present implementation and should be considered in spectral interpretation.

The final waveform is the sum of phonatory, oral-noise, and aspiration components. The radiated kernels include the oral and, where connected, nasal contributions of the tract. Mouth and nostril radiation use smoothed time derivatives of outlet volume flow with adjustable gains. A weak additional wall-radiation term is driven by low-pass acoustic pressure near the glottis; it contributes to the radiated signal without feeding back into the tract or supplying wall compliance to the storage model. The linearized radiation convention includes a reference observation distance of $1\,\mathrm{m}$, but the additional playback and source gains mean that an exported waveform is not an absolute sound-pressure calibration.

This separation of source construction, response calculation, and waveform assembly allows the same utterance to be inspected as a complete signal or as individual source contributions. For a reproducible comparison, these contributions must be rendered with the same settings and common amplitude scaling, rather than normalized independently. Their inspection is particularly useful for distinguishing a short oral release source from a subsequent glottal aspiration interval.

\section{Illustrative Synthesis Examples}
\label{sec:examples}
The three German words in \hyperref[fig:tasse-aerodynamics]{Figures~\ref*{fig:tasse-aerodynamics}}--\ref{fig:lade-aerodynamics} illustrate how articulatory control, aerodynamic state, and acoustic excitation interact within an utterance. They provide complementary contexts for vowels, nasals, laterals, fricatives, and plosives without requiring separate figures for each sound class. The discussion follows the synchronized control, source, pressure--flow, and output panels already introduced above. The vertical reference lines must be distinguished from the temporal organization of articulation. \emph{Acoustically defined segment boundaries} are estimated by inspecting the oscillogram together with the spectrogram (sonagram), using visible changes in the acoustic signal to locate transitions. Such estimates depend on the available acoustic cues; they do not identify universal instants at which all articulators switch from one segment to the next. In the three synthetic examples presented here, the vertical lines are the prescribed segment-timeline boundaries used to organize the intended sound sequence, rather than boundaries independently estimated from acoustic measurements. They provide reference times against which the generated acoustic transitions can be inspected.

Articulatory gestures instead have their own onset, activation build-up, target phase, and decay. These intervals differ across the vocalic, consonantal, velar, and glottal tiers and can overlap across a segment reference line. An acoustic transition reflects the combined consequences of those gestures and the aerodynamic and acoustic state; it need not coincide with any single gesture onset or offset. The pale green activation envelopes and dotted outlines in the upper tiers show the model's actual activation trajectories, including their onset and offset ramps, on a reduced display scale. These coordinated envelopes expose the timing of gestural overlap and the withdrawal of ordinary oral velum support during dominant obstruent or nasal gestures. The vertical reference lines remain shared across the panels to support comparison, not to impose a segmental partition on articulation.

\subsection{Vowels and transitions}
All three words have first-syllable stress and end in unstressed /\textipa{@}/. The short /a/ in ``Tasse'' can be compared with the long /a:/ in ``Lade'', while ``Muse'' supplies the contrasting rounded vowel /u:/. The vocalic tiers show the associated height, frontness, and rounding trajectories. During the vowel intervals, repeated glottal-flow pulses and a predominantly phonatory radiated signal accompany the evolving vocal-tract configuration. The transitions into schwa show the return toward a less differentiated vocalic target while phonation continues. The different vowel durations are specified by the gestural timing; these figures illustrate their realization rather than establish a general model of vowel quantity. The temporal plots show source activity and waveform structure, but do not by themselves quantify vowel formants or spectral accuracy.

\subsection{Nasal and lateral contexts}
The initial /m/ in ``Muse'' (\hyperref[fig:muse-aerodynamics]{Figure~\ref*{fig:muse-aerodynamics}}) combines labial closure with velum lowering and phonation. Opening the nasal branch permits outflow and a radiated signal despite closure of the oral outlet. The subsequent transition to /u:/ combines labial release, a raised velum, and the rounded vocalic configuration. This example illustrates why oral closure alone must not be equated with cessation of glottal flow or acoustic output.

The initial /l/ in ``Lade'' (\hyperref[fig:lade-aerodynamics]{Figure~\ref*{fig:lade-aerodynamics}}) supplies a complementary voiced context. Apical constriction coexists with an effective lateral passage, represented by the bypass-area correction described above. The phonatory component continues into /a:/ without an intervening complete oral stop. This realization illustrates the use of an effective lateral opening within the tube model; it does not resolve the two lateral channels individually.

\subsection{Voiceless and voiced frication}
The medial /s/ in ``Tasse'' (\hyperref[fig:tasse-aerodynamics]{Figure~\ref*{fig:tasse-aerodynamics}}) combines apical narrowing with a glottal opening gesture. The source panel shows oral friction while the phonatory component is suppressed. The /z/ in ``Muse'' (\hyperref[fig:muse-aerodynamics]{Figure~\ref*{fig:muse-aerodynamics}}) instead combines oral friction with continuing phonatory excitation. Its constriction also changes the pressure and glottal-flow traces. These two examples illustrate the distinction between a predominantly noisy excitation and simultaneous periodic and noisy excitation under voiced constriction.

The source curves are RMS amplitudes of separately synthesized components, with the same fixed display multipliers in every figure: voicing $\times1$, oral friction $\times6$, and aspiration $\times5$. Thus the visibility of /z/ friction does not depend on normalizing it to its own maximum. Conversely, the small residual oral-noise component in ``Lade'' remains close to the baseline even with the same magnification. The RMS panel summarizes source strength over $5$-ms windows; it is not a cycle-resolved demonstration of voiced-fricative modulation or a measurement of the noise spectrum.

\subsection{Plosive closure, release, and aspiration}
The initial /t/ in ``Tasse'' (\hyperref[fig:tasse-aerodynamics]{Figure~\ref*{fig:tasse-aerodynamics}}) illustrates the coordination of oral closure and pulmonary pressure onset. The initial closing gesture is advanced by $25$ ms, retaining its release time, so the tract is closed before pressure rises. During closure, intraoral pressure accumulates while oral outflow is blocked. Opening permits a short oral-noise contribution, followed by glottal aspiration and the emergence of vowel phonation. The separately displayed source components distinguish these contributions even where they overlap in the radiated waveform.

The medial /d/ in ``Lade'' (\hyperref[fig:lade-aerodynamics]{Figure~\ref*{fig:lade-aerodynamics}}) provides a voiced-plosive context between vowels. Apical closure produces pressure accumulation and a marked reduction of glottal flow and radiated phonatory amplitude, despite the continuing phonatory control gesture. Release restores flow and vowel excitation. No substantial aspiration component appears in this example. The contrast with /t/ therefore concerns the coordination of glottal control, pressure, and oral release, rather than merely the presence of a closure marker.

Together, the examples demonstrate the operation of the implemented synthesis and the interpretability of its intermediate quantities. They are model illustrations, not matched natural--synthetic comparisons or a perceptual validation. A detailed study of glottal pulse shape and voice quality is outside the scope of these examples.

\section{Discussion and Limitations}
\subsection{Acoustic quality and evaluation status}
The acoustic quality of the present synthesis has not yet undergone a formal evaluation. Source parameters and articulatory--acoustic settings have been adjusted by the author of this paper through iterative listening and inspection of model outputs. The primary practical objective has been to make the intended speech sounds recognizable across the implemented sound classes; high perceptual naturalness has not been the overriding optimization criterion. Recognizability is therefore a development objective and an informal observation, not a result established by independent listeners or a controlled identification experiment.

In the author's listening assessment, vowel quality is already very good, and nasal and lateral realizations are also satisfactory. Fricative noise remains the principal area for further acoustic refinement: the current realizations provide useful consonantal distinctions, while their spectral balance and perceived naturalness would benefit from additional calibration and systematic comparison with natural speech. These judgments describe the present development status and should not be interpreted as comparative quality scores. Independent sound-identification tests and naturalness ratings, complemented by acoustic analyses across speakers and contexts, are needed to determine how well these impressions generalize. Detailed refinement of glottal pulse shape and voice quality is reserved for separate work.

\subsection{A controllable vocal tract for developmental learning models}
A central intended use of DYNARTmo is to serve as the controllable vocal-tract \emph{plant} of a neural model of speech acquisition. In control terminology, the plant is the system acted upon by a controller: here, articulatory commands change the vocal-tract configuration, which determines aerodynamic conditions and acoustic output. It is distinct from an internal forward model that a learner might acquire to predict those consequences. DYNARTmo supplies an explicit articulatory and acoustic environment for such learning; the present implementation does not claim to reproduce individual muscle mechanics or to constitute a learning brain model itself.

This division has a precedent in DIVA, where a neural controller acts on an articulatory synthesizer. During an initial babbling phase, exploratory movements generate paired motor and sensory experience; subsequent imitation learning uses heard speech targets to acquire sound-specific control \citep{tourville2011diva}. For DYNARTmo, the intended progression is likewise from exploratory articulatory activity through imitation toward early word production. The resulting command--configuration--sound trajectories could supply training experience for a neural model, allowing sensorimotor knowledge to emerge from the consequences of its own actions. Coupling to an auditory learner, defining appropriate representations of articulatory state or contact, and implementing the developmental learning stages remain future work.

The purpose of this plant is therefore broader than producing a convincing waveform for a prescribed word. It provides an inspectable relation between a movement, the resulting tract geometry, pressure and flow, and the generated sound. A future learner could explore which actions produce stable acoustic outcomes and which combinations support sound contrasts and syllable sequences. A controllable production system is a necessary component of this particular closed-loop modeling approach; the present paper does not claim that the existing examples already demonstrate acquisition or that every model of language development requires this specific architecture.

\subsection{Articulatory constraints and early speech development}
Developmental observations motivate treating the production apparatus as an active constraint on learning. In a study of nine-month-old infants with unrepaired cleft palate, \citet{chapman2001vocal} reported less canonical babbling and smaller consonant inventories than in age-matched infants without clefts. These findings concern early vocal and consonantal development, rather than a uniform slowing of all language abilities. A later study found smaller expressive lexicons at 21 and 27 months and a greater proportion of words beginning with sonorants in toddlers with cleft palate; the authors considered both lexical selection and difficulties identifying produced obstruents as possible explanations \citep{hardinjones2014lexical}.

For the proposed modeling program, these observations support the hypothesis that the articulatory outcomes available to a child help shape the sensorimotor experience from which speech is learned. They do not isolate anatomy as the sole cause of developmental differences or imply a general impairment of neural learning. An explicit production plant offers a way to investigate such hypotheses by varying articulatory or aerodynamic constraints while specifying the learning conditions. Such experiments would require dedicated anatomical modifications, developmental data, and validation; cleft-palate simulation is not an established capability of the model presented here.

\subsection{Scope of the physical approximation}
The present architecture makes source contributions and their articulatory conditions accessible to inspection, but it retains the approximations described above. Acoustic responses use frame-specific tract geometries, glottal motion is prescribed, and source spectra and amplitudes include empirical calibration. Oral-noise drive uses summarized pressure and glottal flow rather than a resolved local turbulent jet. Its duration follows the evolving aerodynamic drive in an open tract, without an additional constriction-area gate. These properties define the current plant and the experience it would provide to a learner. Improved acoustic calibration and independent evaluation are therefore important both for synthesis quality and for interpreting future developmental simulations.

\section{Conclusions and Further Work}
DYNARTmo combines articulatory movement generation with aerodynamic calculation and acoustic synthesis in an inspectable production model. The examples illustrate how vocal-tract configuration and glottal control give rise to complementary phonatory, friction, and aspiration components. Its principal contribution is a usable articulatory plant controlled through physiologically interpretable movement variables, rather than acoustic output parameters alone. This provides a useful interface for future neural controllers operating at an articulatory or gestural level, including hypotheses about premotor representations, without claiming a one-to-one correspondence between model variables and cortical activity.

Further work will improve perceptual naturalness and physical realism, supported by independent evaluation. A priority is to derive more of the source behavior directly from geometry, pressure, and flow, reducing auxiliary and sound-dependent amplitude corrections wherever a better-founded relation can replace them. Future neural-learning applications will use not only acoustic output but also model-derived sensory information: contact location, onset, and duration at the lips and tongue--palate interfaces, together with articulator-position signals such as jaw height as proprioceptive proxies. Contact strength would require an additional mechanical model or an explicitly identified proxy. These extensions would allow the articulatory plant to supply complementary auditory and somatosensory experience for developmental learning studies.

\section*{Supplementary Material and Reproducibility}
The ancillary material supplied with this article contains version 0.1.0 of a standalone Python reference implementation of the single-geometry acoustic waveguide and branched mean-flow models described here. It includes source code, explicit parameters and units, illustrative geometries, model limitations, and numerical validation fixtures exported from the Dart implementation underlying Speech Articulation Trainer 5.2.0. Nine aerodynamic and fifteen acoustic cases, comprising 255,048 scalar comparisons, agree within an absolute tolerance of $10^{-12}$ or a relative tolerance of $10^{-9}$, whichever is larger. These comparisons establish numerical agreement for the tested cases, not independent physical or perceptual validation. The package permits noncommercial use, modification, and redistribution under its included licence. The application interface, gesture orchestration, fitting workflow, calibrated noise-source generators, and utterance-level overlap-and-add procedure are not included; the Python package therefore does not reproduce complete spoken words.

A short accompanying guide explains how to generate the three German word examples directly from the application's synthesis rules, without loading a natural recording or importing a saved timeline. In Sound~$\rightarrow$~syll, the respective inputs are \verb|'ta.s@#| (``Tasse''), \verb|'mu:.z@#| (``Muse''), and \verb|'la:.d@#| (``Lade''). The initial apostrophe marks stress, the period separates syllables, the colon lengthens the preceding segment, \verb|@| denotes schwa, and the final \verb|#| selects statement intonation. After confirming the input with Enter, users enable synthesis under Values~$\rightarrow$~Set and calculate and play the result under Sound~$\rightarrow$~move. In this release, the +Synthesis and +Natural buttons are coupled; enabling them does not require loading or using a natural reference.

The guide describes the rule-based workflow rather than promising sample-identical reproduction of the figures. In particular, the ``Tasse'' figure additionally advances the beginning of the initial /t/ target phase by 25~ms while retaining its release time; this timing edit is not encoded in the transcription. Natural recordings, reference-aligned segment timelines, gesture-score files, and synthetic word WAV files are not included in the supplement. Reproduction of the model components is distinguished from word-level demonstration in the application and from future imitation or fitting experiments. The ancillary material includes the Python source package, a short word-example guide, and a plain-text transcription list. Speech Articulation Trainer 5.2.0 is available at \url{https://speech-articulation-trainer.web.app}. It initially presents Articulation mode; +Synthesis and +Natural can be enabled together under Values~$\rightarrow$~Set. The versioned reference package identifies the precise Dart baseline through source hashes and includes instructions for repeating the numerical comparisons. The article's CC BY 4.0 licence does not apply to the separately licensed Python package; its noncommercial source licence is included with that package.

\section*{Author Contributions and AI Assistance}
The author established the scientific objectives and articulatory framework and contributed to the conception, methodological development, and interpretation of the synthesis model. Specific contributions included the coordination and dominance of articulatory gestures, their activation build-up and decay, and the distinction between control frames, glottal pulse times, and acoustic response duration. The author also contributed to the conceptual development of overlapping acoustic responses for changing vocal-tract configurations, including consideration of tract-length variation. He critically examined the aerodynamic assumptions and the plausibility of calculated pressure and airflow behavior, identified modeling problems through inspection of the intermediate quantities, and directed subsequent revisions. This expert scrutiny is distinct from independent experimental validation. The author provided the listening examples, assessed successive synthesis revisions, and defined the intended use of the model in future neural simulations of speech acquisition.

Development proceeded through an iterative dialogue between the author and OpenAI's ChatGPT/Codex tools. These tools were used extensively to generate and revise program code, implement and test model changes, prepare figures, and assist with drafting and revising the manuscript. A substantial part of the programming was carried out through this AI-assisted workflow. The author's scientific contributions included specifying and refining the modeling requirements, evaluating proposed solutions, and guiding their revision; the AI assistance extended to implementation and manuscript preparation. AI-generated code, analyses, and text remain subject to the author's review and verification. The author retains responsibility for the final software and manuscript.

\clearpage

\begin{figure}[p]
\centering
\definecolor{Exampleheight}{HTML}{1687a7}
\definecolor{Examplefront}{HTML}{be6068}
\definecolor{Exampleround}{HTML}{516b86}
\definecolor{Examplelab}{HTML}{1687a7}
\definecolor{Exampletip}{HTML}{327bc2}
\definecolor{Exampledors}{HTML}{777777}
\definecolor{Examplevel}{HTML}{21843b}
\definecolor{Examplega}{HTML}{be6068}
\definecolor{Examplelp}{HTML}{1687a7}
\definecolor{Examplect}{HTML}{be6068}
\definecolor{Exampleasp}{HTML}{9560b5}
\definecolor{Examplephon}{HTML}{005533}
\definecolor{Exampleoral}{HTML}{d78216}
\definecolor{Exampleag}{HTML}{a12aa8}
\definecolor{Exampleug}{HTML}{21843b}
\definecolor{Exampleps}{HTML}{a75b17}
\definecolor{Examplept}{HTML}{ed8b23}
\definecolor{Examplewave}{HTML}{333333}
\resizebox{\linewidth}{!}{
}
\caption{Lade /\textipa{\textprimstress la:d@}/. The five upper tiers show vocalic, consonantal, velar, glottal, and pulmonary/CT control. Pale green envelopes with dotted outlines show the actual gesture activations, including onset and offset ramps; their heights are scaled for display. In the vocalic and velum tiers, activation zero coincides with trajectory zero. Positive velum values indicate raising/closing and negative values lowering/opening. The vertical lines mark segment-timeline reference boundaries, not gesture boundaries. The shared amplitude panel shows voicing, oral friction, and aspiration, as the absolute RMS amplitudes of their separately synthesized radiated components in non-overlapping 5-ms windows. For display only, voicing is multiplied by 1, oral friction by 6, and aspiration by 5. These fixed factors, amplitude limits, and synthesis gains are identical across all three words; no per-source or per-word peak normalization is applied. These are output amplitudes in arbitrary units, not calibrated sound pressures. Area and flow are reconstructed from responses associated with glottal pulse times; pressures use successive control frames. The final panel shows the radiated floating-point synthetic output before PCM quantization.}
\label{fig:lade-aerodynamics}
\end{figure}

\begin{figure}[p]
\centering
\definecolor{Exampleheight}{HTML}{1687a7}
\definecolor{Examplefront}{HTML}{be6068}
\definecolor{Exampleround}{HTML}{516b86}
\definecolor{Examplelab}{HTML}{1687a7}
\definecolor{Exampletip}{HTML}{327bc2}
\definecolor{Exampledors}{HTML}{777777}
\definecolor{Examplevel}{HTML}{21843b}
\definecolor{Examplega}{HTML}{be6068}
\definecolor{Examplelp}{HTML}{1687a7}
\definecolor{Examplect}{HTML}{be6068}
\definecolor{Exampleasp}{HTML}{9560b5}
\definecolor{Examplephon}{HTML}{005533}
\definecolor{Exampleoral}{HTML}{d78216}
\definecolor{Exampleag}{HTML}{a12aa8}
\definecolor{Exampleug}{HTML}{21843b}
\definecolor{Exampleps}{HTML}{a75b17}
\definecolor{Examplept}{HTML}{ed8b23}
\definecolor{Examplewave}{HTML}{333333}
\resizebox{\linewidth}{!}{
}
\caption{Muse /\textipa{\textprimstress mu:z@}/. The five upper tiers show vocalic, consonantal, velar, glottal, and pulmonary/CT control. Pale green envelopes with dotted outlines show the actual gesture activations, including onset and offset ramps; their heights are scaled for display. In the vocalic and velum tiers, activation zero coincides with trajectory zero. Positive velum values indicate raising/closing and negative values lowering/opening. The vertical lines mark segment-timeline reference boundaries, not gesture boundaries. The shared amplitude panel shows voicing, oral friction, and aspiration, as the absolute RMS amplitudes of their separately synthesized radiated components in non-overlapping 5-ms windows. For display only, voicing is multiplied by 1, oral friction by 6, and aspiration by 5. These fixed factors, amplitude limits, and synthesis gains are identical across all three words; no per-source or per-word peak normalization is applied. These are output amplitudes in arbitrary units, not calibrated sound pressures. Area and flow are reconstructed from responses associated with glottal pulse times; pressures use successive control frames. The final panel shows the radiated floating-point synthetic output before PCM quantization.}
\label{fig:muse-aerodynamics}
\end{figure}
\clearpage
\end{document}